\documentclass[11pt,a4paper]{article}
\usepackage{orcidlink}
\UseRawInputEncoding

\usepackage{amsmath}
\usepackage{amssymb}
\usepackage{amsthm}
\usepackage{booktabs}
\usepackage{multirow}
\usepackage{authblk} 
\usepackage{multicol}
\usepackage{longtable}
\usepackage[numbers]{natbib}
\usepackage{float}
\usepackage{bm}
\usepackage{enumerate}
\usepackage{graphicx}
\usepackage{lipsum}
\usepackage{array}
\usepackage{colortbl}
\usepackage[all]{xy}
\usepackage{tikz}
\usepackage{verbatim}
\usepackage[left=2cm,right=2cm,top=3cm,bottom=2.5cm]{geometry}
\usepackage{hyperref}
\usepackage{caption}
\usepackage{subcaption}
\usepackage{psfrag}

\title{Bridging Gaps in Natural Language Processing for Yor\`{u}b\'{a}: A Systematic Review of a Decade of Progress and Prospects}

\author[1]{Toheeb A. Jimoh\thanks{Corresponding author: toheeb.jimoh@ul.ie}\orcidlink{0000-0003-3830-7641}}
\author[1]{Tabea De Wille \orcidlink{0000-0001-8575-6162}}
\author[1]{Nikola S. Nikolov \orcidlink{0000-0001-8022-0297}}
\affil[1]{Department of Computer Science and Information Systems, University of Limerick, Castletroy, V94 T9PX, Limerick, Ireland}
\date{} 
\begin{document}
\maketitle
\begin{abstract}
\noindent
    Natural Language Processing (NLP) is becoming a dominant subset of artificial intelligence, as the need for machines to understand human language becomes increasingly indispensable. Several NLP applications are ubiquitous, partly due to the myriad datasets being churned out daily through mediums like social networking sites. However, the growing development has not been evident in most African languages due to persistent resource limitations, among other issues.  Yor\`{u}b\'{a} language, a tonal and morphologically rich African language, suffers a similar fate, resulting in limited NLP usage. To encourage further research towards improving this situation, this systematic literature review aims to comprehensively analyse studies addressing NLP development for Yor\`{u}b\'{a}, identifying challenges, resources, techniques, and applications. A well-defined search string from a structured protocol was employed to search, select, and analyse 105 primary studies between $2014$ and $2024$ from reputable databases. The review highlights the scarcity of annotated corpora, the limited availability of pre-trained language models, and linguistic challenges like tonal complexity and diacritic dependency as significant obstacles. It also revealed the prominent techniques, including rule-based methods, statistical methods, deep learning, and transfer learning, which were implemented alongside datasets of Yor\`{u}b\'{a} speech corpora, among others. The findings reveal a growing body of multilingual and monolingual resources, even though the field is constrained by socio-cultural factors such as code-switching and the desertion of language for digital usage. This review synthesises existing research, providing a foundation for advancing  NLP for Yor\`{u}b\'{a} and in African languages generally. It aims to guide future research by identifying gaps and opportunities, thereby contributing to the broader inclusion of Yor\`{u}b\'{a} and other under-resourced African languages in global NLP advancements.  \\

\end{abstract}

\newpage
\section{Introduction}
Natural Language Processing (NLP) is one of the key components of artificial intelligence. It has rapidly gained prominence in recent years as the need to help machines understand human languages grows. NLP involves developing intelligent systems that can converse with humans through natural language \cite{thanaki2017python}. These intelligent systems are products of the two main classes of NLP: natural language understanding and natural language generation. These classes coalesce into the generation of language and understanding its intricacies, respectively \citep{Yeldo2021NaturalLP}. \\

\noindent
Natural language typically denotes languages spoken and used by humans via various communication channels for day-to-day interactions. In contrast, artificial languages, like computer languages, are those created with certain rules and restrictions \citep{dipanjan}. Moreover, unlike artificial language, natural language possesses ambiguity, making its processing often a hard nut to crack. For instance,
the statement, ``Erik Ten Hag wins his $50^{\text{th}}$ game with Manchester United," poses a situation where
there could be two distinct meanings, depending on if one is saying Erik won his $50^{\text{th}}$ game in charge of the club, or that Erik won his $50^{\text{th}}$ career game with Manchester United. NLP applications have been seen in various exciting domains such as sentiment analysis \cite{muhammad2023_afrisenti, shode2023_nollysenti, muhammad2022_naijasenti}, machine translation \cite{adebara2022_linguistically_motivated, ayogu2018_statistical_MT, eludiora2016_eng2yor_MT, ahia2021low-resource_double_bind}, named entity recognition \cite{ayogu-ner, adelani2021_masakhaner_1, mehari2024semi-supervised_NER, adelani2022_masakhaner_2}, parts-of-speech (POS) tagging \cite{Abiola_hidden_markov, enikuomehin_implementation_lagos_women, dione2023_masakha_pos}, question answering \cite{ogundepo-etal-2023-cross}, amongst others, aiming to bridge the communication gap between humans and
machines. \\

\noindent
NLP techniques have witnessed fascinating evolution with diverse languages worldwide since their inception in the 1950s \cite{nlp_evolution_1950}, initially with the rule-based approach. This involves defining linguistic rules for core language concepts, including semantics, pragmatics, morphology and phonology. These techniques have metamorphosed from the rule-based approach to statistical approaches, subsequently incorporating machine learning techniques, which utilise large-scale linguistic resources. Undoubtedly, the explosion of large datasets, mainly through social networking sites, necessitated the implementation of more advanced techniques. This gives rise to the introduction of deep learning methods \cite{lecun2015_deep_learning}, built on neural networks. Interestingly, the field of NLP witnessed revolutionisation through the advancement in deep learning, with notable models like the recurrent neural networks (RNNs), long short-term memory (LSTM), and more recently, the transformer architectures \citep{evolution_of_NLP}.
Moreover, this development has improved NLP performance in machine translation and question-answering tasks. More importantly, it gives way to a better understanding of sequential dependencies in natural language \citep{Tripathi2018NaturalLP_dependecies}. \\

\noindent
Furthermore, self-attention mechanisms and bidirectional training in models such as bidirectional encoder representations from transformers (BERT), alongside the generative pre-trained transformer (GPT-3) model developed by OpenAI\footnote{\url{https://openai.com/}}, resulted in remarkable improvement in natural language understanding \citep{Kulkarni2023TheEO_LLM}. This innovation continues to set the pace for the emergence of many improved large language models (LLMs) like the Large Language Model Architecture (LLaMA) \cite{Touvron2023_llama}, Mistral \cite{Jiang2023_Mistral}, and others. LLMs, usually trained on huge datasets, have been recorded to have achieved state-of-the-art result performance across various tasks, which transitions from task-specific to task-independent architectures \cite{Patil2024ARO_llm_performance}. However, the huge dataset needs of these models translate into a \textit{boon and bane} for NLP in the context of many under-resourced languages in the world. These languages, because they are less digitised, sparingly taught, harbour resource scarcity \& low density, are less privileged, among other identifiers, are referred to as low-resource languages (LRLs) \citep{singh-2008-natural_LRLs, adeboje_bilingual_transformer}. \\

\noindent
Despite these significant advancements in NLP for major global languages like English and Chinese, underrepresented languages like Yor\`{u}b\'{a}---a language spoken by about 50 million people \cite{fagbolu2016_applying_rough_set, adewole2020_vowel_elision} primarily in Nigeria and its diaspora---remain under-explored in computational linguistics research. Yor\`{u}b\'{a}'s linguistic richness, characterized by tone marking, complex morphology, and extensive oral traditions, presents unique challenges and opportunities for NLP development. Recent studies have underscored the importance of developing NLP resources and techniques for low-resource languages to ensure equitable access to technology, thereby conserving linguistic diversity. For Yor\`{u}b\'{a}, these early efforts include tasks such as diacritic restoration \cite{orife_sequence2sequence}, machine translation \cite{eludiora2016_eng2yor_MT}, sentiment analysis \cite{eludiora2015_word_sense_disambiguation}, and parts-of-speech tagging \cite{ayogu2017comparative}. However, the dearth of annotated datasets, tools, and computational models tailored to Yoruba significantly hampers progress in the field.\\

\noindent
To promote NLP research involving Yor\`{u}b\'{a} language, it is essential to examine the current landscape of research efforts, identify existing gaps, and evaluate available tools and resources. While some work has been done in isolated tasks such as diacritic restoration, machine translation, and sentiment analysis, there is no comprehensive review that brings together these fragmented efforts. As a result, researchers and developers lack a centralised knowledge base to inform further progress in Yor\`{u}b\'{a} NLP. This lack of synthesis impedes collaboration, resource-sharing, and efficient advancement in the field. This study, therefore, aims to fill this critical gap by conducting a systematic literature review (SLR) that synthesises existing research in NLP for the Yor\`{u}b\'{a} language. Specifically, the review seeks to: \begin{itemize}
    \item Identify the range of NLP tasks addressed in the literature,
    \item Categorise the methods and techniques employed,
    \item Document the existing resources and tools, and 
    \item Highlight the unique linguistic challenges and how they have been addressed.
\end{itemize} 
These insights will offer a clear map of what has been achieved so far, where the limitations lie, and where future research should be directed. The guiding research questions for this review are explicitly outlined in Section~\ref{sec3.2}.\\

\noindent
The intended contribution of this review is twofold. Firstly, it serves as a foundational reference for researchers, developers, and language technologists interested in Yor\`{u}b\'{a} and other low-resource African languages, by offering a structured overview of existing work and identifying key gaps and opportunities. Secondly, by revealing progress and bottlenecks, this review can support targeted development of language tools, datasets, and models tailored to the linguistic features of Yor\`{u}b\'{a}. In doing so, it contributes to the broader goals of linguistic preservation, technological inclusion, and equitable representation of African languages in NLP, which is an effort supporting the United Nations' Sustainable Development Goals (SDGs). Ultimately, the outcomes of this review are expected to guide future research agendas, inform policy and funding decisions, and foster collaborations toward inclusive NLP development.

\section{Background of study}
This section briefly introduces the Yor\`{u}b\'{a} language by discussing its constituents, such as letters and types, and detailing its roots regarding its language family.  
\subsection{Yor\`{u}b\'{a} language overview}
Yor\`{u}b\'{a} language is one of the largest low-resource African languages with over $47$ million speakers, encompassing several dialects with considerable similarities \cite{ahia2024voices_unheard, olaleye2023_yfacc}. It is adopted as a native and social language in Western African countries, including Nigeria, Togo, Benin Republic, and other countries like Cuba, Brazil, etc. \citep{developing_yoruba_corpus}. Yor\`{u}b\'{a} uses $25$ out of the $26$ Latin-script letters, excluding \textit{q, z, v, x} and \textit{c} \cite{akinade2023_varepsilon}. Thus, an additional $4$ letters---\textit{\d{e}, \d{o}, \d{gb}} and \textit{\d{s}}---with the existing $21$ makes up the entirety of the Yor\`{u}b\'{a} alphabets. In addition, it is a tonal language with three primary lexical tones:  high, medium, and low \cite{akinade2023_varepsilon}. The tones are usually represented by acute (as in \'{u}),  grave (as in \`{u}), and an optional macron (as in \={u}), denoting the high, low and mid-tone, respectively \cite{adelani2021_masakhaner_1}. The three tonal signs and the underdots cater for diacritics, which determine the linguistic meaning of words in the language. Generally, the language is composed of $7$ vowels (\textit{a, e, \d{e}, i, o, \d{o}, and u}), about $5$ nasal vowels, (\textit{an, \d{e}n, in, \d{o}n, and un}), and $18$ consonants (\textit{b, d, f, g, gb, h, j, k, l, m, n, p, r, s, \d{s}, t, w, and y}) \cite{okediya2019_building_ontology, orife_sequence2sequence}.\\

\noindent
As derived by the Expanded Graded Intergenerational Disruption Scale (EGIDS) categorisation \cite{simons2010making_EGIDS}, the language currently has the institutional level status. This implies that it has achieved sizable development and is still utilised beyond the community and individual homes. This is evident in its usage and availability of resources through written books, mass media, and various undocumented oral traditions.  However, it is still classified as a low-resource language alongside other Nigerian languages due to the dearth of basic computing resources \citep{Asubiaro2021ASR}. This indicates that available resources remain untapped for creating natural language corpora and developing technological and NLP tools \cite{ishola2020_yoruba_dependency_treebank}.   \\

\subsection{Yor\`{u}b\'{a} Language Family}
\begin{figure}[htbp]
\centering 

\includegraphics[width=1.0\textwidth]{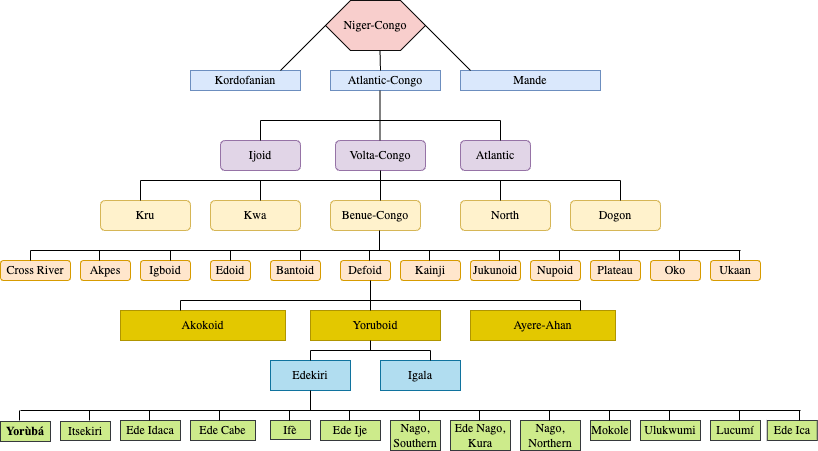}
\caption{Yor\`{u}b\'{a} Language Family Tree}
\label{fig1} 
\end{figure}

The Yor\`{u}b\'{a} language belongs to the Niger-Congo family \cite{ishola2020_yoruba_dependency_treebank}. This language family is the most prominent and largest of the four major African linguistic groups: Niger-Congo, Nilo-Saharan, Afro-Asiatic and Khoisan \citep{BendorSamuel2024}. This language family possesses distinctive noun class systems; nonetheless, they exhibit substantial variations in types, especially in morphological complexity \citep{Good2018NigerCongoL}. A substantial part of languages of sub-Saharan Africa---containing Western Africa, Southern Africa, Eastern Africa, and Central Africa---belong to this family, making up about $85\%$ of the entire African language population \citep{mussandi2024nlp_african}.  Among these are Cape Town, South Africa, in the southern part; Dakar, Senegal, in the western part; and Mombasa, Kenya, in the east of Africa. Figure  \ref{fig1} presents a language tree visualisation showing the connection of these language groups or families. It starts with the Niger-Congo family and shows the groups and subgroups up to the last node, displaying the position of the Yor\`{u}b\'{a} language, including others in the same class.\\

\noindent
At the time of writing this paper, the Niger-Congo language family had 1552 language subgroups, which are classified into three main groups: Atlantic-Congo, Mande, and Kordofanian\footnote{\url{https://www.ethnologue.com/subgroup/47/}}. In addition to three groups, a language group, \textit{Mbre}, is left unclassified within the Niger-Congo language family. The \textit{Kordofanian} branch is seen to be primarily spoken by the Nuba people of southern Sudan. Unlike the former, the \textit{Mande} languages are commonly spoken in Western African countries, mostly in The Gambia, Burkina Faso, Senegal, and Mali, to mention a few. Similarly, the \textit{Atlantic-Congo} languages are mostly used in a similar demographic as the \textit{Mande} languages. The languages are used mainly in Western African countries like Liberia, Guinea, Guinea-Bissau, Senegal, The Gambia, and Sierra Leone. Moreover, they are documented as the most prominent, diverse, and primary component of the Nigeria-Congo language family \cite{merrill2021_atlantic_congo}. \\

\noindent
Going down the language tree, it is seen that the Yor\`{u}b\'{a} language stems from the Atlantic-Congo language group, recognised as the most used in the Niger-Congo language family. Moreover, the \textit{Ijoid}, \textit{Atlantic} and the \textit{Volta-Congo} language groups emerge from this \textit{Atlantic-Congo} group. Among these three branches, the Yoruba follows the path of the \textit{Volta-Congo} subgroup, which was recorded to contain $5$ branches: \textit{Kwa}, \textit{Kru}, \textit{Dogon}, \textit{North} and the \textit{Benue-Congo} subgroups. From these, about $12$ language family subgroups are documented to be contained in the \textit{Benue-Congo} subgroups, which house the Yor\`{u}b\'{a} language. However, one language, \textit{Fali of Baissa}, is left unclassified. Furthermore, the \textit{Defoid} subgroup, from the \textit{Benue-Congo} languages begets the \textit{Yoruboid} group, which in turn contains the \textit{Edekiri} and \textit{Igala} language subgroups. Consequently, the \textit{Yor\`{u}b\'{a}} language stems from this \textit{Edekiri} branch. Other language in the same class as the \textit{Yor\`{u}b\'{a}} language include: \textit{Itsekiri}, \textit{Ede Idaca}, \textit{Ede Cabe}, \textit{If\`{e}}, \textit{Nago, Southern}, \textit{Ede Ije}, \textit{Ede Nago, Kura}, \textit{Nago, Northern}, \textit{Mokole}, \textit{Ulukwumi (Ol\`{u}k\`{u}mi)}, \textit{Ede Ica}, and \textit{Lucum\'{i}}. In summary, the 
\textit{Yor\`{u}b\'{a}} language belongs to the \textit{Niger-Congo} language family, sequentially from \textit{Atlantic-Congo}, \textit{Volta-Congo}, \textit{Benue-Congo}, \textit{Defoid}, \textit{Yoruboid}, with \textit{Edekiri} as its last branch.


\section{Materials and Methods}
This section details the review processes, from planning to data synthesis. The guidelines comprehensively detailed in \cite{kitchenham2007guidelines, page2021prisma} were followed to ensure adequacy in the review processes, from the selection of studies to the reporting stage. These guidelines help identify, analyse, and interpret all information from the primary studies considered without bias or prejudice. 

\subsection{Review Planning}\label{sec3.1}
A systematic literature review requires formal planning and meticulous attention to detail. Consequently, a defined protocol was followed to ensure the success of the entire process. To minimise the manual rigour typically required in systematic reviews, three key tools were employed: Covidence\footnote{\url{https://www.covidence.org/}}, Zotero\footnote{\url{https://www.zotero.org/}}, and Microsoft Excel. Covidence streamlined multiple stages of the review process. It enabled the bulk import of studies in RIS format and automatically detected and removed duplicate records from various article databases, eliminating the need for manual cross-checking. Covidence also facilitated the title/abstract and full-text screening phases through a structured interface that supported inclusion/exclusion tagging and decision logging by multiple reviewers. Zotero served as the reference management platform, supporting efficient bibliographic organisation, duplicate checking, and citation formatting. Microsoft Excel was used to define a structured data extraction template and to streamline the quality assessment phase, thereby reducing the likelihood of manual errors. \\

\noindent
These tools collectively reduced human effort in screening, deduplication, tagging, and data organisation. Furthermore, they addressed methodological bottlenecks, including inconsistent inclusion decisions, manual duplicate removal, error-prone data tracking, and fragmented reviewer collaboration. Ultimately, they helped in improving reproducibility, transparency, and efficiency, as well as the adoption of a more reliable and scalable review workflow. \\
 
 \begin{figure*}[htbp]
\centering 
\includegraphics[width=1.0\textwidth]{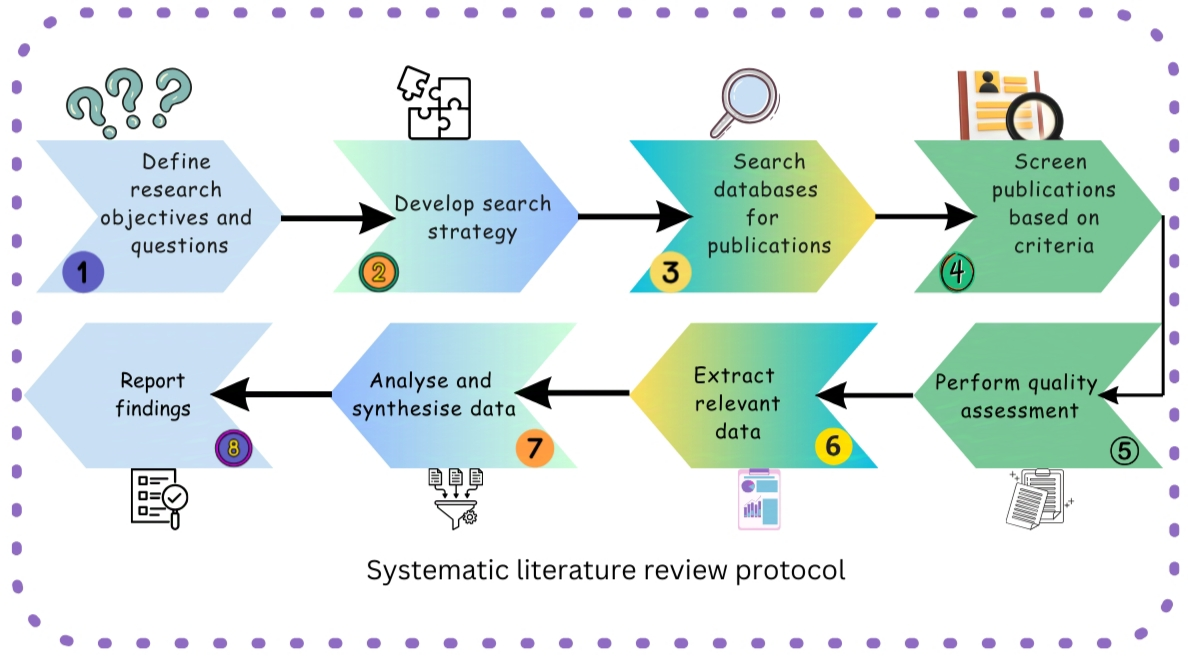}
\caption{The systematic literature review protocol}
\label{fig2} 
\end{figure*}

\noindent
Moreover, following the review protocol shown in Figure \ref{fig2}, the first task was to define the research objectives and develop the research questions based on the objectives. Thereafter, the search strategy was developed. This includes determining the search string and the scope of the search. Subsequently, potential papers are searched through identified databases and screened based on the selection criteria. A quality assessment is performed on the included paper to ensure standard review and that the relevant data are extracted thereafter. Furthermore, the available data are analysed to obtain the results reported comprehensively afterward. The following sections provide a more detailed explanation of the processes.

\subsection{Research Questions}\label{sec3.2}
Research questions were developed to define a precise template for the broad study objectives. Four research questions presented in Table~\ref{tab1} were carefully formulated to guide a structured and focused review of the current state of NLP research involving the Yor\`{u}b\'{a} language. Each question targets a distinct dimension of the NLP pipeline and contributes to achieving the study’s overall objective: to investigate the development, resources, methods, and challenges in Yor\`{u}b\'{a} NLP. RQ1 helps to identify the variety of computational tasks that have received attention in Yor\`{u}b\'{a} NLP research. It is meant to establish the breadth of work done and uncover underexplored tasks. RQ2 focuses on the computational methods and frameworks used across these studies. This helps to track methodological trends and possibly evaluate how well existing approaches align with the linguistic complexities of Yor\`{u}b\'{a}. Additionally, RQ3 aims to assess the existing resources and ultimately identify resource gaps that hinder progress. Finally, RQ4 addresses the core limitations and bottlenecks that researchers face, helping to understand the challenges that are vital for proposing actionable solutions and advancing future research in various domains. \\

\noindent
Collectively, these research questions seek to provide a well-encompassing basis for analysing the current state of Yor\`{u}b\'{a} NLP, highlighting strengths, gaps, and priorities for further investigation, directly supporting the main objectives of the study.

\begin{table}[htbp]
    \centering
    \caption{Research questions}
    \begin{tabular}{lp{7cm}}
    \hline
    \hline
       $RQ_i$  & Research Questions  \\
       \hline
       \hline
    
        $RQ_1$ & What NLP tasks have been addressed for Yor\`{u}b\'{a} language? \\
        $RQ_2$ & What techniques have been employed for Yor\`{u}b\'{a} NLP? \\
        $RQ_3$ & What language resources are available for Yor\`{u}b\'{a} language? \\
        
        $RQ_4$ & What challenges are associated with NLP development involving Yor\`{u}b\'{a} language? \\
        \hline
    \end{tabular}
    \label{tab1}
\end{table}

\subsection{Search Strategy}\label{sec3.3}
This section describes the gathering of the vital primary studies and the \textit{systematic} steps toward achieving the goal. This is a crucial phase designed to eliminate potential bias and incorporate randomization in the selection and sample size determination. Reputable databases related to the subject matter were thoroughly examined to identify all relevant studies for the systematic review, utilizing a well-defined search strategy. Generally, the central goal of the strategy involves attracting studies that have applied computational NLP approaches involving Yor\`{u}b\'{a} language, be it as a monolingual, bilingual, or multilingual NLP set-up.

\subsubsection{Method and Scope of Search}
The strategy initially employed an automated search method to obtain studies relevant to the systematic review. This method involved digging into each electronic database with the defined keywords, boolean operators, and wild cards---as and when due. Furthermore, to ensure a representative sample and a higher recall, $9$ databases were targeted in total, viz: 
    Web of Science\footnote{\url{https://www.webofscience.com/}}, ScienceDirect\footnote{\url{https://www.sciencedirect.com/}}, Google Scholar\footnote{\url{https://scholar.google.com}}, Association for Computing Machinery (ACM) Digital Library\footnote{\url{https://dl.acm.org/}}, Institute of Electrical and Electronics Engineers (IEEE) Xplore\footnote{\url{https://ieeexplore.ieee.org/}}, Semantic Scholar\footnote{\url{https://www.semanticscholar.org/}}, Scopus\footnote{\url{https://www.scopus.com/}}, ScienceDirect\footnote{\url{https://www.sciencedirect.com/}}, and Association for Computational Linguistics (ACL) Anthology\footnote{\url{https://aclanthology.org/}}. \\

\begin{figure}[htbp]
\centering 
\includegraphics[width=0.9\textwidth]{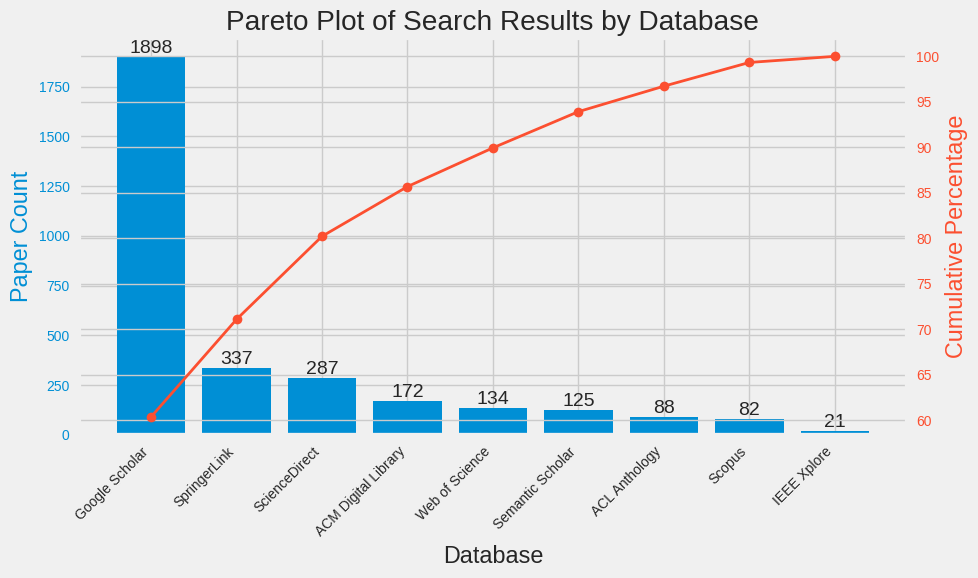}
\caption{Initial Search Results across Databases}
\label{fig3} 
\end{figure}

\noindent
Moreover, with the motive of exploring a decade of the progress of NLP research in the Yor\`{u}b\'{a} language paradigm, studies between  $2014$ and $2024$ were chosen. This decade range was also established by considering the time of writing this paper, and it is plausible since the studies of $2014$ were published towards the year's end. In addition, to ensure potent quality, only peer-reviewed journal articles and conference papers were included in the search results, with the language of writing exclusively in English. Additionally, due to the computational requirements of NLP research, studies primarily in Computer Science, Engineering, and Computational Linguistics were considered, with a strict focus on empirical studies. Figure \ref{fig3} is a Pareto plot used to show the statistics of the initial search results across the databases, alongside the cumulative percentage. It shows that most papers were from Google Scholar, accounting for about $60 \%$  of the search results, while the least were from IEEE Xplore. The curve shows the cumulative percentage of studies from each database. It indicates the percentage contribution of each database, with the least being IEEE Xplore.

\subsubsection{Search Strings} 
The search strings denote the combination of keywords used in exploring the databases to extract potential primary studies. Potential keywords in the systematic review objectives were identified and combined to build this. However, before the combination, a few related research publications were scanned through to see the usage of the relevant phrases and words. Thus, since the study seeks to synthesize information vis-\`{a}-vis NLP research and solutions associated with Yor\`{u}b\'{a} language, the general search string is formulated based on the three segments and presented as follows:

\textit{(``Natural Language Processing" OR NLP OR ``Computational linguistics" OR ``neural network*" OR statistic* OR ``machine learning" OR ``artificial intelligence" OR corpus OR dataset* OR ``data set*") AND Yoruba AND (solution OR task OR application)
} \\

\noindent
However, it is pertinent to mention that the search string was modified slightly for different databases based on their unique requirements and search functionalities, and the domains were restricted to computer science and computational linguistics. Moreover, to obtain search records from Google Scholar, a software tool named ``Publish or Perish'', which obtains and analyzes scholarly citations, was used. This tool leverages many data sources, including Google Scholar, to get the raw citations seamlessly. \\

\subsubsection{Study Selection Criteria}\label{sec3.3.3}
Having obtained a substantial number of studies considered relevant to the study, the criteria to painstakingly select the most essential and directly aligned studies were defined. This involves a set of defined statements based on the research objectives and scope, making up the inclusion and exclusion criteria. It accounts for the desired year range, publication type, access to full-text documents, etc. Table \ref{tab2} shows the inclusion and exclusion criteria, which are defined following the PICO criteria: population, intervention, context, and outcome \cite{kitchenham2007guidelines}. In addition, \textit{Study Range} and \textit{Publication Type}, and \textit{Publication Language} were used to represent the publication years considered for the primary, the type of publication, and the language of publication, respectively. Furthermore, the two classes of criteria---inclusion and exclusion---defined statements to determine if the particular study is to be part of the review or irrelevant to it, respectively. Relevant articles were searched and gathered via the listed databases, bearing in mind the defined criteria. \\

\begin{table*}[htbp]
    \centering
    \caption{Inclusion and exclusion criteria}
    \begin{tabular}{lp{6cm}p{6cm}}
    \hline
    \hline
       S/N  & Inclusion & Exclusion  \\
       \hline
       \hline
        Population &  Studies specifically addressing NLP solutions, tasks, or methods for Yor\`{u}b\'{a}  language & Studies not involving Yor\`{u}b\'{a} language \\
        Comparison & Comparative studies of NLP solutions for low-resource languages, including  Yor\`{u}b\'{a} & Comparative studies not including Yor\`{u}b\'{a}  \\ 
        Outcome & Studies reporting the performance of NLP solutions involving Yor\`{u}b\'{a} & Theoretical studies without practical evaluations or results \\
        Intervention & Study involving NLP solutions for Yor\`{u}b\'{a} language & Study involving NLP solutions for unrelated languages\\
        Study Range & Publications between $2014$ \& $2024$ inclusive & Publications before $2014$ \\
        Publication type & Peer-review journal articles and conference papers & Non-peer review studies, including thesis, reviews, surveys, etc.\\

        Publication language & Studies published in English language & Studies published in languages other than English \\
        
        \hline
    \end{tabular}
    \label{tab2}
\end{table*}

\noindent
Initially, $3144$ studies were obtained across the $9$ databases. The RIS files containing these studies from their respective sources were uploaded to the Covidence software tool. Due to the inevitable intersection across databases, the software detected a total duplicate count of $816$, while $50$ duplicates were manually detected. Again, following the selection criteria, abstract screening was initially carried out, leaving out a total of $1865$ from the unique $2278$ studies. Moreover, full-text screening was carried out on the remaining $413$ studies, and a total of $308$ studies were discarded based on defined reasons. Eventually, $105$ primary studies were obtained for inclusion in the review.\\

\begin{figure*}[htbp]
\centering 
\includegraphics[width=0.9\textwidth]{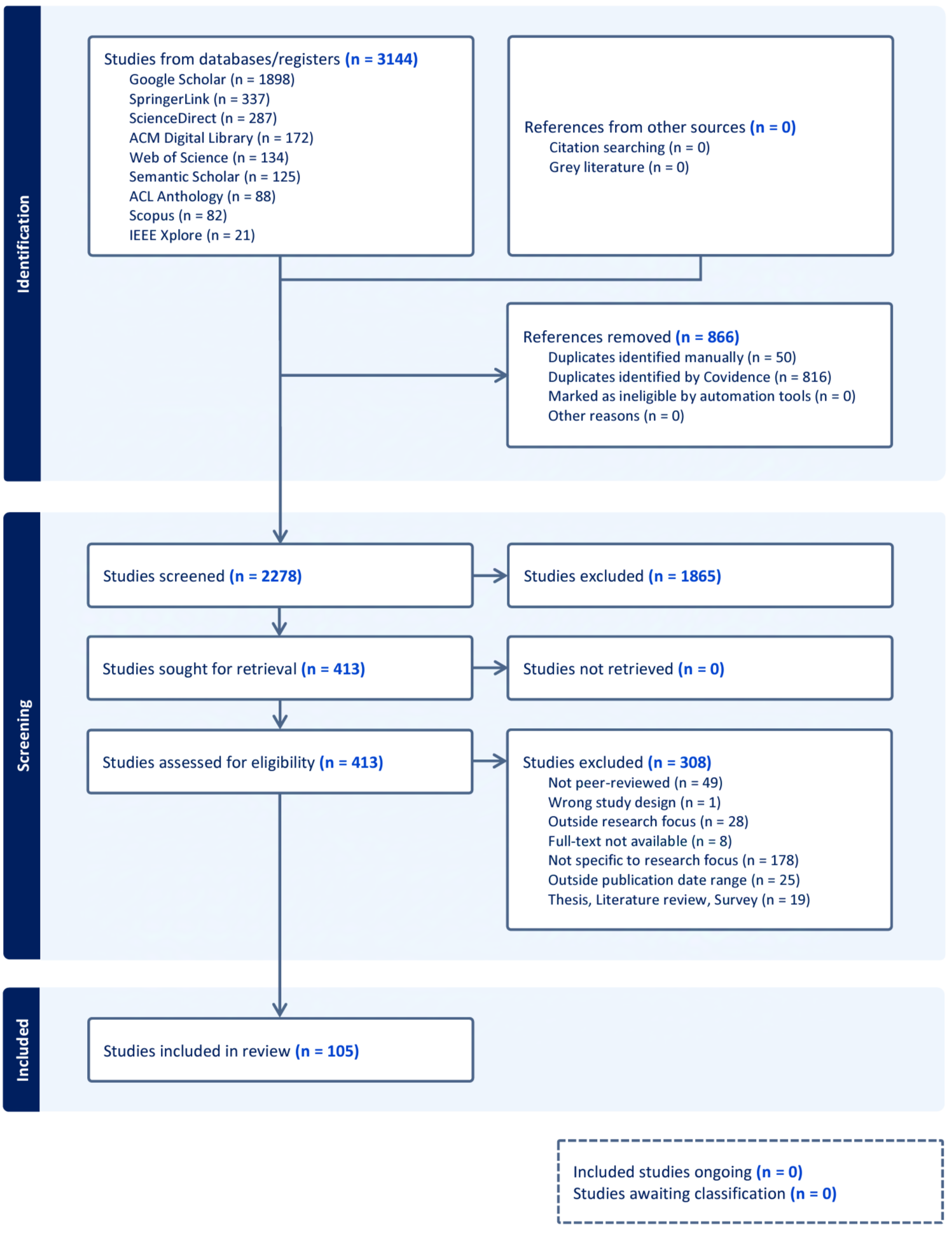}
\caption{PRISMA diagram of selection process}
\label{fig9} 
\end{figure*}

\noindent
Subsequently, backwards and forward snowballing techniques were also used to capture studies that might have been missed in database searches due to conflicting terminology, following \cite{wohlin2014_snowballing_guidelines}. For backward snowballing, reference lists of a quarter of the $105$ articles were surveyed to check for additional relevant studies. It was observed that the identified publications were part of the initial search results across databases. Additionally, forward snowballing, which involves checking for publications that have cited the potential primary study, was carried out. Here, $5$ pre-prints and $1$ review publications were discovered; however, they were not included due to the defined exclusion criteria. The snowballing loops were ended after no new studies were obtained. Ultimately, $105$ primary studies were included in the SLR after conducting a quality assessment check, presented in Section \ref{sec3.4}. The whole selection process ensures a detailed and comprehensive overview by incorporating the Preferred Reporting Items for Systematic Reviews and Meta-Analysis (PRISMA) framework, which is also presented in Figure \ref{fig9}, following \cite{page2021prisma}.

\subsection{Quality Assessment}\label{sec3.4}
This section shows further efforts to ensure the credibility of primary studies used in the systematic literature review. Specific quality questions were used for verification to further ascertain the reliability of the primary studies to be eventually included in the SLR. Thus, a quality checklist was created, following the guidelines in \cite{kitchenham2007guidelines}. The response associated with each question is defined through a multiple-choice response: Yes, Partially, and No. A  ``Yes" response shows higher certainty, thus getting a score of $2$ . Similarly, a ``Partially" response is assigned $1$, while a ``No" response is assigned $0$. For this research purpose, a total quality score below the median score is to be excluded. The median is calculated as the value in the $\Big(\frac{n+1}{2}\Big) ^{\text{th}}$ position, where $n$ is the maximum quality score, which equals $16$. Thus, for quality scores $S_i = \{1, 2, \dots, 16\}$, the median position, $M_p$, is derived as follows: 
\begin{align}
    M_p &= \left(\frac{n+1}{2}\right)^{th} \text{position}
    \end{align}
\begin{align*}
   M_p &= \left(\frac{16+1}{2}\right)^{th} \text{position}  \\
    \therefore M_p&= 8 \frac{1}{2}  
\end{align*}
Thus, the median, $M$, is 
\begin{align*}
    M=\frac{8+9}{2}=8.5
\end{align*}

\noindent
Consequently, a paper with a score less than $8.5$ out of the $16$ total score will be excluded. However, no study falls in this range. Hence, the total $105$ primary studies are included in the systematic review. The quality assessment technique is based on the study by \cite{van2022predictive} and was extended using a measure of central tendency, the median, as it accurately measures the center of a dataset's distribution and is also not affected by outliers. Table \ref{tab3} shows the checklist containing the quality domain and the questions asked for each. \\

\begin{figure}[htbp]
\centering 
\includegraphics[width=0.9\textwidth]{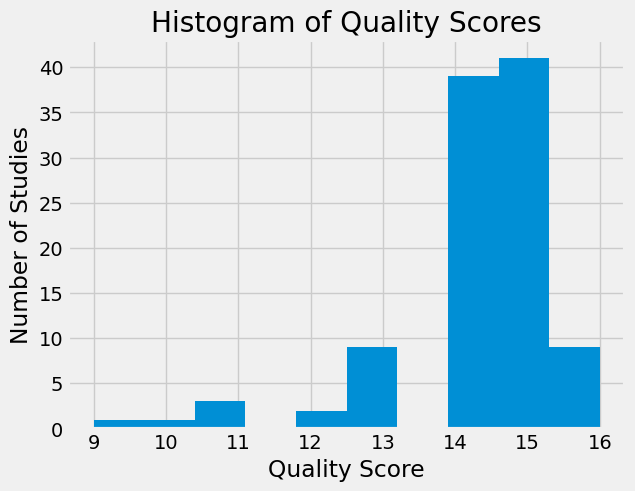}
\caption{Histogram of total quality score for each study}
\label{fig4} 
\end{figure}

\begin{table}[htbp]
    \centering
    \caption{Quality Assessment Checklist}
    \begin{tabular}{lp{10cm}}
    \hline
    \hline
       Domain  & Question  \\
       \hline
       \hline
        Research objective & Are the objectives of the study clearly stated?\\
        Methodology \& study design & Are the study's methods and experimental design clearly defined? \\
         Research documentation & Are the study's processes comprehensively documented? \\
        Research Question Alignment & Are the research questions answered through the findings? \\
        Study conclusion &  Do the conclusions of the study relate to its objectives?\\
        Result evaluation & Does the study validate the results using standard evaluation metrics? \\
        Limitations and bias & Are the limitations of the study acknowledged? \\
         Novelty and contribution & Does the study contribute new tools, resources, insights, or investigate new questions? \\
        \hline
    \end{tabular}
    \label{tab3}
\end{table}
\noindent
After assessing each potential primary study based on the quality questions, scores were assigned\textemdash a maximum of 16 for each. The results are plotted through the histogram in Figure \ref{fig4}. The plot shows that most of the primary studies have a score of $15$, followed by $14$. Moreover, only a minute quantity of the primary studies have a total score between $9$ and $12$, with $9$ being the minimum total score being $9$, while the maximum is $16$. This shows high quality in the included primary studies, potentially making this review's results highly valuable.

\subsection{Data Extraction}
Data extraction involves collecting information relevant to the SLR from the primary studies. It was carried out by defining an extraction template. This template is structured based on the following sections: Bibliographic information, research context, NLP focus, language scope, results and evaluation, and contributions. The sections, in turn, contain individual elements that constitute all the extraction items. Pilot extractions were carried out with fewer samples to verify the results and update the form based on the need or irrelevance of variables to be extracted. For this study, the ``research country" and ``research continent" are taken as the affiliation of the lead author at the time of their research publication. The final extraction template is given in Table \ref{tab4}.

\begin{table}[htbp]
    \centering
    \caption{Data Extraction Template}
    \begin{tabular}{lp{5cm}}
    \hline
    \hline
       Variable  & Extraction Item  \\
       \hline
       \hline
        V1 & Study ID\\
        V2 & Article Title \\
        V3 & Publication Year \\
        V4 & Research Country \\
        V5 & Research Continent \\
        V6 & Publication Type\\
        V7 & Publication Source \\
        V8 & NLP Task \\
        V9  & NLP Technique\\
        V10 & Language Scope \\
        V11 & Dataset \\
        V12 & Citation Count \\
        \hline
    \end{tabular}
    \label{tab4}
\end{table}

\subsection{Data Synthesis}
This section involves integrating and interpreting information extracted from the selected studies. Initially, data were obtained from individual databases and combined to form the primary studies after excluding required studies based on the selection criteria. Findings were obtained from the primary studies across diverse domains, including the NLP tasks, techniques, resources, and challenges. Moreover, the process was carried out using Microsoft Excel and the Covidence platform, mentioned explicitly in Section \ref{sec3.1}.  The ultimate goal at this stage is to identify patterns, trends, and knowledge gaps to inform future research directions in Yor\`{u}b\'{a} NLP.

\section{Results and Interpretations}
This section discusses the results obtained from the primary studies included in the systematic review through the data synthesis. It helps answer the research questions and provides visual illustrations of every component of the primary studies.
\subsection{General Statistics}
\begin{figure}[htbp]
\centering
\begin{minipage}{0.48\textwidth}
    \centering
    \includegraphics[width=\linewidth]{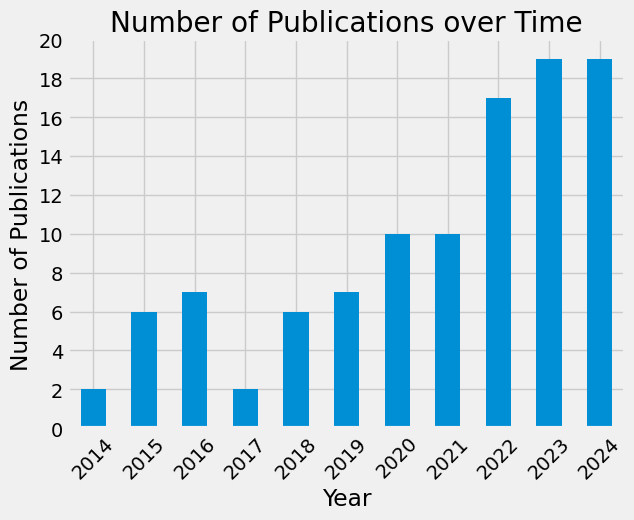}
    \caption{Distribution of publications per year}
    \label{fig5}
\end{minipage}\hfill
\begin{minipage}{0.5\textwidth}
    \centering
    \includegraphics[width=\linewidth]{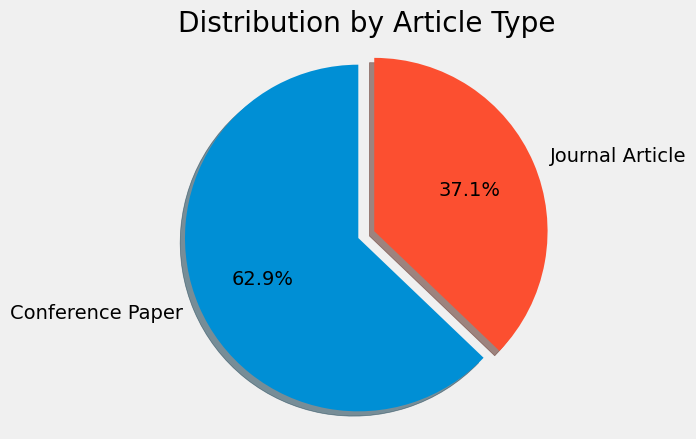}
    \caption{Distribution of publication types}
    \label{fig6}
\end{minipage}
\end{figure}

The primary studies included in the research, totalling $105$, were synthesized for relevant data. They consist only of journal articles and conference papers based on the selection criteria specified in the systematic review protocol. Figure \ref{fig5} reveals the frequency of the sample\textemdash included primary studies\textemdash over the years. It attests that the Yor\`{u}b\'{a} NLP research has grown over the years, with an apparent upward trend after the break in year $2017$. In addition, $2023$ and $2024$ have the highest number of primary studies, even though the latter year is yet to end at the time of this review.
Subsequently, the publications are assessed based on the type. Figure \ref{fig6} shows the distribution of article type, in which there are more conference papers $(62.9\%)$ than journal articles $(37.1\% )$. This observation is plausible as most researchers in this domain are primarily involved in conferences, probably as conference publications are typically shorter and faster to publish. \\

\begin{figure}[htbp]
\centering 
\includegraphics[width=0.65\textwidth]{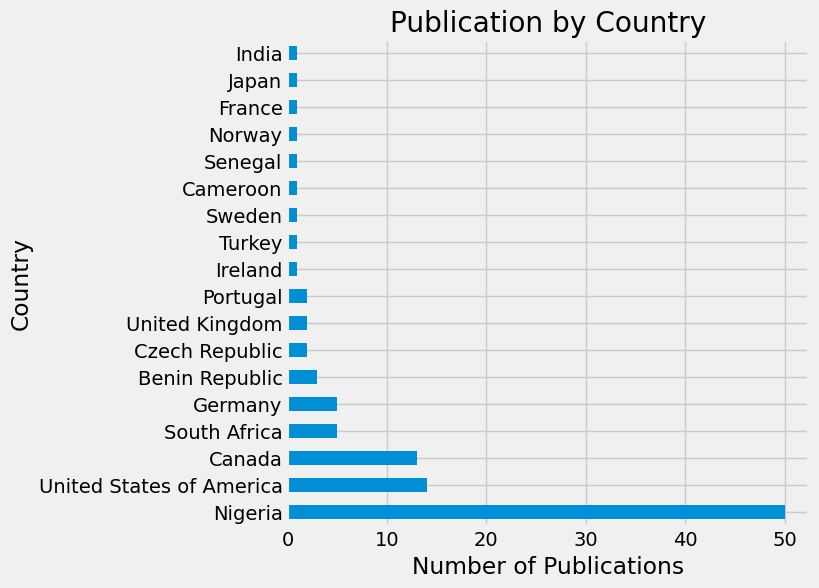}
\caption{Authors' geographical location}
\label{fig10} 
\end{figure}

\noindent
Furthermore, Figure \ref{fig10} shows the first authors' geographical locations at the publication's time. It reveals an interesting insight, making it known that most publishers were based in Nigeria at the time of publishing, while a considerable number were also in the United States, Canada, South Africa, Germany, and the Benin Republic. This tells that research in this domain extends beyond the dominant speaking country of the Language---Nigeria. It also shows promising outcomes due to visible collaborations taking place among researchers across the world. \\

\begin{figure}[htbp]
\centering 

\includegraphics[width=.6\textwidth]{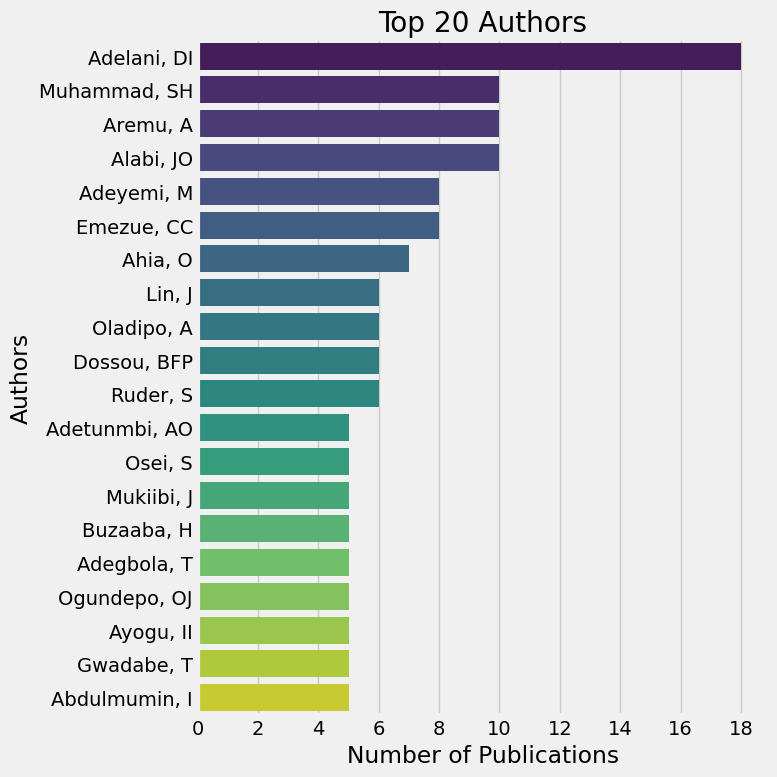}
\caption{Popular authors by publication number}
\label{fig7} 
\end{figure}

\noindent
Finally, accessing the authors' contributions to this field shows an informative trend, as depicted in Figure \ref{fig7}. It shows the top 20 authors in this field based on the primary studies. The top author, David Adelani, has about $18$ publications in total. Moreover, there have been observable collaborations among the top $7$ authors, including Dossou, BFP; and Osei, S. Similarly, significant collaborations exist among Lin, J; Oladipo, A; and Ogundepo, OJ, among other notable collaborations. This diagram reveals famous authors in the research domain.

\subsection{Research Question Analysis}
This section contains pertinent responses to specific research questions outlined by the systematic literature review objectives. The analysis presented herein represents a key contribution to the review. It summarises the inquiry findings through structured responses to each question, making it an essential tool for answering similar questions for future researchers.

\subsubsection{RQ1: What NLP tasks have been addressed for Yor\`{u}b\'{a} language?}

Investigating the various NLP tasks involved with applying Yor\`{u}b\'{a} language has revealed improved and promising results over the years. These NLP tasks or solutions synthesis considered the application of the language either as a monolingual NLP research or as part of a bilingual or multilingual task---only considered when there is a vivid emphasis on the language. \\

\noindent
Efforts have been made to democratize the numbering system of Yor\`{u}b\'{a} language through NLP, with earlier studies by making digital content and technologies more inclusive and usable for native speakers. \cite{akinade2014_number-to-text} tried to develop a computational system for converting cardinal numbers into their Yor\`{u}b\'{a} number names in a number-to-text conversion task, enabling machine-readable and voice-accessible numerical expressions in Yor`{u}b'{a}. Similarly, \cite{elesemoyo_numeral} develops a machine translation system that translates English number expressions into Yor\`{u}b\'{a} text, taking into account the linguistic and syntactic context in which the numbers appear.  While the former focused on the text normalization task, a core foundational NLP task required in this research domain, the latter focused on machine translation. Moreover, diacritic restoration, an important component of text normalization tasks, has been given considerable research efforts, as seen from \cite{orife_sequence2sequence, ayogu2021_diacritic, alqahtani2019_efficient_CNN, alqahtani2019investigating_input-output_diacritic} studies. Ultimately, three studies \cite{oluwaseyi2024_spelling, oladiipo2020_spelling_error_patterns, asahiah2023_diacritic} have made efforts toward improving spell-checking and correction in the Yor\`{u}b\'{a} texts. These tasks are crucial for accurately representing texts in the language, thereby improving the availability of quality data. \\

\noindent
The need to understand sentence structure due to its importance for the syntactic parsing of text has also made parts-of-speech tagging essential research towards improving Yor\`{u}b\'{a} NLP. This involves assigning grammatical categories, such as nouns, adjectives, adverbs, etc., to words based on the meaning or context. Four studies \cite{dione2023_masakha_pos, Abiola_hidden_markov, ayogu2017comparative, enikuomehin_implementation_lagos_women} worked on this task, whereas \cite{dione2023_masakha_pos} involved $20$ typologically diverse African languages, in which Yor\`{u}b\'{a} was a key subset of. Similarly, NLP tasks involving morphological induction \cite{adegbola2022_automatic_morphology} and analysis \cite{agbeyangi2016_morphological_standard_yoruba_nouns, adegbola2016_pattern-based}  were explored as these are essential for breaking words into their roots and affixes. In addition, two studies \cite{asahiah2021_comparison, akinwonmi2024_rule-based_misanalysis} investigated syllabification tasks, which involve breaking words into their syllables. These tasks are essential for ensuring precise semantic and syntactic parsing in texts. \\

\noindent
Furthermore, identifying entities, such as names of people, organizations and places, in texts is crucial to advancing Yor\`{u}b\'{a} NLP. This involves the named entity recognition aspect of NLP and has been investigated in a significant number of studies, including \cite{ayogu-ner, beukman2023_ner, adelani2021_masakhaner_1, adelani2022_masakhaner_2, mehari2024semi-supervised_NER}. These studies are essential for identifying specific entities, such as cultural components and personal names in Yor\`{u}b\'{a}, and are, in turn, vital for downstream tasks.  A study \cite{dione2021_multilingual_wolof} also developed a dependency parsing multi-task model in a multilingual approach by considering Wolof, Bambara, and Yor\`{u}b\'{a} language as core components.  The study is crucial for analyzing grammatical relationships among words in a sentence, and it involves syntactic knowledge transfer from high-resource to the extremely low-resource languages referenced. \\

\noindent
Many studies also explored the crucial task involving automatic translation of text data between Yor\`{u}b\'{a} and other languages. About $18$ studies explored the machine translation (MT) tasks, cutting across rule-based machine translation (RBMT) \cite{adewole2020_vowel_elision, safiriyu2015_personal_pronouns, Adegoke_Elijah_2023_xml_encoded, I__2015_tone_changing, babatunde2021english_android_phones, elesemoyo_numeral, eludiora2016_eng2yor_MT}, statistical machine translation (SMT) \cite{ayogu2018_statistical_MT, babatunde2024_speech-to-text_hybrid}, neural machine translation (NMT) \cite{adeboje_bilingual_transformer, ahia2021low-resource_double_bind, Esan_2020_rnn, liu2018_context_OOV, adebara-etal-2024-cheetah, adelani2021effect_domain_and_diacritic}, and a few hybrid MT methods involving SMT and RBMT \cite{fagbolu2016_applying_rough_set}; RBMT and example-based machine translation (EBMT) \cite{Adewale_real-time_chatting}; SMT and NMT \cite{adebara2022_linguistically_motivated}. Moreover, a similar task on sentence alignment, which is crucial to NMT, was explored in \cite{signoroni2023_evaluating_sentence_alignment}, considering a bilingual approach involving the English-Yor\`{u}b\'{a} pair.\\

\noindent
Language modelling is also an indispensable task, as it is often necessary to predict the likelihood of word sequences to capture linguistic patterns in languages. Only a handful of studies have been seen to work in this domain; the related studies of \cite{dossou-etal-2022-afrolm, oladipo2023_better} specifically involve multilingual language model pre-training, which were developed for African languages, with significant emphasis on  Yor\`{u}b\'{a} language. Moreover, to verify the effectiveness of a modest amount of data for multilingual language modelling, \cite{ogueji2021_small-data} introduced the AfriBERTa model, while \cite{jude-ogundepo-etal-2022-afriteva} also introduced AfriTeVa, to further extend language modelling with limited training data to sequence-to-sequence modelling.  Furthermore, the semantic modelling task was explored in \cite{okediya2019_building_ontology}. It involves the construction of a Yor\`{u}b\'{a} language ontology meant to characterise the semantic relationships between the words, among which are antonyms, synonyms, hyponyms, hypernyms, holonyms and meronyms. \\

\noindent
Understanding sentiment-bearing phrases or words in text and speech is equally essential to Yor\`{u}b\'{a} NLP for determining the language users' emotional undertone. Research in sentiment analysis has also been explored across a few studies, with most using a multilingual approach. \cite{raychawdhary2024_optimizing_multilingual_SA, raychawdhary2023transformer} strictly investigate sentiment analysis while \cite{shode2023_nollysenti, muhammad2022_naijasenti, Adeniji_disambiguation_framework, muhammad2023_afrisenti} includes language resource development alongside the sentiment analysis tasks addressed. In addition, research which solely involves text \cite{Adegoke_Elijah_2023_xml_encoded, ayogu2020exploring_naive_bayes, ogueji2021_small-data} or topic classification \cite{adelani2023_masakhanews, hedderich2020_transfer_learning_distant} was also explored in five studies altogether. Moreover, under these broader classification categories, the language identification or detection task was also carried out in two studies \cite{adebara-etal-2022-afrolid, asubiaro2018word}. \\

\noindent
Even though most studies tend to involve text processing, quite a substantial number of studies also investigated speech processing in Yor\`{u}b\'{a}. This ranges from text-to-speech synthesis \cite{isewon_grapheme_based, boco2022_end2end, aoga2016integration_maryTTS, abdulkareem2016_yorcall}, tone recognition for continuous speech \cite{bengono2024improving_tone_recognition}, speech recognition \cite{isaac2023_deep_reinforcement, sosimi2019standard_svm, rahmon2024_speech_recognition}, text-to-speech analysis \cite{iyanda2015_statistical_zipf}, speech-based gender recognition \cite{sefara2019_yoruba_gender_recognition_neural_network} and multi-tasks involving both text-to-speech and speech-to-text \cite{Akintola_hci}, text-to-speech and language speech resource development \cite{bibletts, dagba2016_design_tts, developing_yoruba_corpus, antenatal, ogunremi2024_iroyinspeech}, speech synthesis and language pitch modelling \cite{van2014predicting_utterance_pitch_target}, and two studies exclusively focusing on speech corpus development \cite{sosimi2015supervised, Akinwonm2021_prosodic}. Whereas, \cite{olaleye2023_yfacc} differs as it is based on corpus development for visually grounded speech. Furthermore, to optimize modelling in speech recognition, acoustic unit discovery tasks \cite{ondel2022non_bayesian, ajayi2022acoustic, yusuf_hierarchical} are essential in learning speech sounds embedding to retain the linguistically relevant acoustic information and discard the irrelevant ones. \\

\noindent
Ultimately, increased usage of social networking sites and websites, among others, has undoubtedly increased the extent of information available in recent years. Thus, developing effective methods of accessing and retrieving information is indispensable. Few studies also explored this, including tasks on dense information retrieval \cite{oladipo2024_backbones}, monolingual retrieval tasks \cite{zhang2023_miracl}, and cross-lingual information retrieval tasks \cite{ogundepo-etal-2023-cross, onifade2018_embedded_fuzzy_dictionary, adeyemi2024_ciral}. \\

\noindent
These insights show considerable progress with the involvement of Yor\`{u}b\'{a} language in various NLP tasks, culminating in invaluable efforts at solution development in the research domain. Table \ref{tab5} is used to summarize the various NLP tasks involving Yor\`{u}b\'{a} as presented by the primary studies. 

	



\subsubsection{RQ2: What techniques have been employed for \textit{Yor\`{u}b\'{a}} NLP? }
\noindent
The metamorphosis of NLP techniques involving Yor\`{u}b\'{a} language is observed to not necessarily lag behind the general trend of accomplishment in the NLP community. Most of the earlier research works are seen to be limited to implementing rule-based techniques \cite{isewon_grapheme_based, adewole2020_vowel_elision, safiriyu2015_personal_pronouns, oladiipo2020_spelling_error_patterns} before the advancement, leading to incorporating statistical modelling \cite{sosimi2015supervised, asubiaro2018word, ayogu2018_statistical_MT, oluwatoyin_stochastic}, and sometimes, a combination of both techniques \cite{Adeniji_disambiguation_framework}. Furthermore, recent studies have incorporated supervised machine learning \cite{ayogu-ner, ayogu2021_diacritic, ayogu2020exploring_naive_bayes, sosimi2019standard_svm} and unsupervised learning \cite{agbeyangi2020_authorship_verification, yusuf_hierarchical, adegbola2022_automatic_morphology, ondel2022non_bayesian, adegbola2016_pattern-based} techniques in this research domain, and a few times, the hybrid of statistics and machine learning techniques \cite{ayogu2017comparative, oluwatoyin_stochastic}. Lately, deep learning architectures \cite{raychawdhary2023transformer, raychawdhary2024_optimizing_multilingual_SA, alqahtani2019_efficient_CNN, orife_sequence2sequence, adeboje_bilingual_transformer}, as well as a hybrid of machine and deep learning methods \cite{olalekan2022_machine_learning_for_naija}, are being utilised for various tasks as well. \\

\begin{figure}[htbp]
	\centering
	\begin{subfigure}{0.49\textwidth}
		\includegraphics[width=\textwidth]{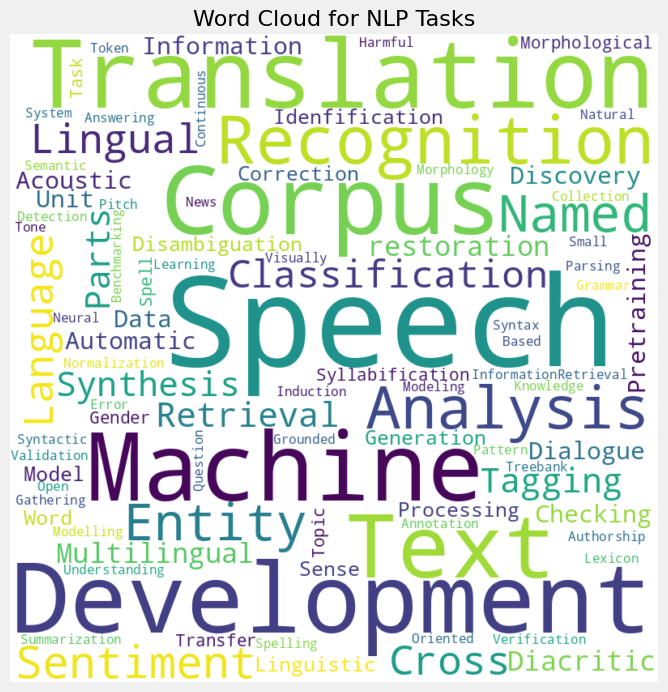}
		\caption{Common NLP tasks}
		\label{fig8a}
	\end{subfigure}
	\begin{subfigure}{0.49\textwidth}
		\includegraphics[width=\textwidth]{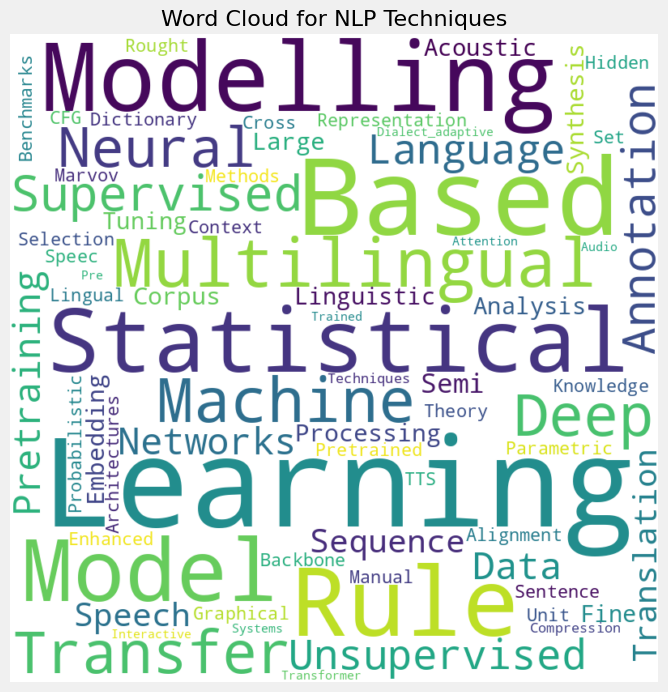}
		\caption{Common NLP techniques}
\label{fig8b}
	\end{subfigure}
	\caption{Word clouds of NLP tasks and techniques}
	\label{fig8}
\end{figure}

\noindent
Also, due to the low-resource nature of Yor\`{u}b\'{a} language, NLP research works in its domain have benefited from transfer learning by pre-training on a high-resource language and fine-tuning on specific tasks in this language. Transfer learning of this nature over the years has been mainly composed of leveraging multilingual pre-training \cite{adebara-etal-2024-cheetah, adebara2023_serengeti, gehrmann2023_tata, ruder2023_xtreme-up, dione2021_multilingual_wolof, ogueji2021_small-data} and cross-lingual transfer \cite{ajayi2024_crosslingual_harmful, ogundepo-etal-2023-cross, dione2023_masakha_pos, adeyemi2024_ciral}. Figure \ref{fig8} shows word clouds of both the employed techniques \ref{fig8b} and the NLP tasks \ref{fig8a} carried out in the primary studies involved in the SLR. \\

\noindent
Table \ref{tab6} shows a cross-tabulation of the specific NLP tasks involving  Yor\`{u}b\'{a} language together with the NLP techniques utilised in each task.  It reveals that only studies involving sentiment analysis and TTS tasks utilised all the techniques as classified. The MT task closely followed the previous two tasks vis-\`{a}-vis the varieties of techniques involved in their studies; all the techniques classes were utilised in the MT studies, except for the machine learning technique. Moreover, other tasks have studies utilising most of the classes of techniques; however, acoustic unit discovery, ASR, spell checking \& correction, speech-to-text, and tone identification \& recognition tasks include studies using only two different techniques. In addition, only the speech-based gender recognition task has studies involving only a single technique---a deep learning technique. 

\subsubsection{RQ3: What language resources are available for Yor\`{u}b\'{a}  language?}

Many low-resource languages experience stunted growth in their language development due to the flaw associated with their lack of substantial datasets required for various NLP tasks amidst recent data-hungry models. This factor has undoubtedly prompted significant efforts at developing various language resources for Yor\`{u}b\'{a} language to further improve the performance of various NLP tasks in the language. The developed resources have been seen to cut across different domains; while some studies solely focused on resource development, others utilised the datasets or corpora on specific tasks. The summary of the corpora, which contains Yor\`{u}b\'{a} either as a monolingual, bilingual, or as an important part of a multilingual dataset, is presented in Table \ref{tab7}. Furthermore, for detailed purposes, specific datasets or corpora with assigned names are briefly outlined individually as follows, while others are listed as part of the general descriptions.\\

\noindent
\textbf{AFRIQA}: AFRIQA \cite{ogundepo-etal-2023-cross} pioneers among cross-lingual question-answering (QA) corpus development for African languages. It is the first cross-lingual open retrieval question-answering (XOR QA) dataset focused on African languages, bridging the gap in resource availability and paving the way for more equitable and inclusive question-answering technologies. More importantly, Yor\`{u}b\'{a} language is a core component among the languages in the dataset, as it comprises about $12,239$ XOR QA examples for $10$ African languages across the southern, western, eastern and central African regions, viz: Bemba, Fon, Hausa, Igbo, Kinyarwanda, Swahili, Twi, Wolof, Yor\`{u}b\'{a} and Zulu. Specifically, for the Yor\`{u}b\'{a} language, the training, development, and testing examples are $360, 261$, and $332$, respectively. Furthermore, experiments to evaluate the performance of state-of-the-art models on the dataset were carried out, and this mainly involved automatic translation, entailing translating questions or retrieved documents in preparation for processing, and multilingual retrieval, involving carrying out direct retrieval through multilingual embeddings.\\

\onecolumn
\begin{longtable}{p{4cm}p{4cm}p{4cm}p{4cm}}
\caption{Summary of NLP Tasks and Corresponding Studies} \\ 

    \hline
    \rowcolor[HTML]{87CEEB} 
    NLP Task & Studies Addressing Task & Popular Techniques & Dataset used  \\ \hline
    \endfirsthead
    \hline
    \rowcolor[HTML]{87CEEB} 
    NLP Task & Studies Addressing the Task & Popular Techniques & Dataset used \\ \hline
    \endhead
    \hline
    \multicolumn{4}{|r|}{\textit{Continued on next page...}} \\ \hline
    \endfoot
    \hline
    \endlastfoot
    Acoustic unit discovery & \cite{ajayi2022acoustic}, \cite{yusuf_hierarchical}, \cite{ondel2022non_bayesian} & Statistical modelling & Yor\`{u}b\'{a} Speech Corpus; Custom data\\ 
    Automatic speech recognition & \cite{ogunremi2024_iroyinspeech}, \cite{sosimi2015supervised}, \cite{rahmon2024_speech_recognition}, \cite{isaac2023_deep_reinforcement} & Statistical modelling & \`{I}r\`{o}y\`{i}nSpeech; Custom data \\ 
    Diacritic restoration & \cite{ayogu2021_diacritic},  \cite{alqahtani2019investigating_input-output_diacritic}, \cite{alqahtani2019_efficient_CNN} & Neural networks &Lagos-NWU Yor\`{u}b\'{a} Speech Corpus; Custom data \\ 
    Language modelling & \cite{okediya2019_building_ontology}, \cite{dossou-etal-2022-afrolm}, \cite{ogueji2021_small-data}, \cite{oladipo2023_better}, \cite{jude-ogundepo-etal-2022-afriteva} & Multilingual pre-training & W\'{U}R\`{A}; Common Crawl Corpus; Custom data \\
    Information retrieval & \cite{onifade2018_embedded_fuzzy_dictionary}, \cite{zhang2023_miracl}, \cite{oladipo2024_backbones}, \cite{adeyemi2024_ciral}, \cite{ogundepo-etal-2023-cross} & Multilingual language modelling & AFRIQA; MIRACL; CIRAL; Custom data \\ 
    Machine translation    & \cite{ayogu2018_statistical_MT}, \cite{fagbolu2016_applying_rough_set}, \cite{adewole2020_vowel_elision}, \cite{adeboje_bilingual_transformer}, \cite{adebara-etal-2024-cheetah}, \cite{eludiora2016_eng2yor_MT}, \cite{safiriyu2015_personal_pronouns}, \cite{liu2018_context_OOV},  \cite{Esan_2020_rnn}, \cite{Adegoke_Elijah_2023_xml_encoded}, \cite{adebara2022_linguistically_motivated}, \cite{Adewale_real-time_chatting}, \cite{I__2015_tone_changing}, \cite{babatunde2021english_android_phones}, \cite{signoroni2023_evaluating_sentence_alignment}, \cite{elesemoyo_numeral}, \cite{babatunde2024_speech-to-text_hybrid}, \cite{adelani2021effect_domain_and_diacritic}, \cite{ahia2021low-resource_double_bind}, \cite{ahia2024voices_unheard}, \cite{vegi2022_webcrawl_african}, \cite{akinade2023_varepsilon}  & Rule-based; statistical modelling; multilingual pre-training &  WebCrawl African;  YOR\`{U}LECT; IkiniYor\`{u}b\'{a}; MENYO-20k; Custom data,  \\ 

    Named Entity Recognition & \cite{ayogu-ner}, \cite{beukman2023_ner}, \cite{adelani2021_masakhaner_1}, \cite{adelani2022_masakhaner_2}, \cite{alabi2020massive_vs_curated}, \cite{mehari2024semi-supervised_NER}, \cite{ogueji2021_small-data}, \cite{hedderich2020_transfer_learning_distant} & Transfer learning; Pre-training multilingual embeddings & MasakhaNER; Common Crawl Corpus; Custom data \\ 
    Part-of-speech tagging & \cite{dione2023_masakha_pos}, \cite{enikuomehin_implementation_lagos_women}, \cite{ayogu2017comparative}, \cite{Abiola_hidden_markov} & Statistical modelling; cross-lingual transfer & MasakhaPOS; Lagos-NWU Yor\`{u}b\'{a} Speech Corpus; Custom data \\ 
    Sentiment analysis & \cite{muhammad2023_afrisenti}, \cite{shode2023_nollysenti}, \cite{muhammad2022_naijasenti}, \cite{Adeniji_disambiguation_framework}, \cite{raychawdhary2023transformer}, \cite{raychawdhary2024_optimizing_multilingual_SA} & Deep learning; transfer learning & AfriSenti; NaijaSenti; NollySenti; Custom data \\
   
    Speech-based gender recognition & \cite{modupe2019_yoruba_gender_recognition_bilstm}, \cite{sefara2019_yoruba_gender_recognition_neural_network} & Neural networks & Lagos-NWU Yor\`{u}b\'{a} Speech Corpus \\ 
    Spell checking \& correction & \cite{oluwaseyi2024_spelling}, \cite{oladiipo2020_spelling_error_patterns}, \cite{asahiah2023_diacritic} & Statistical modelling & Custom data \\ 
    Syllabification & \cite{akinwonmi2024_rule-based_misanalysis}, \cite{asahiah2021_comparison} & Rule-based & Custom data \\ 
    Speech-to-Text & \cite{Akintola_hci}, \cite{ajayi2022acoustic} & Statistical modelling & Yor\`{u}b\'{a} Speech Corpus; Custom data \\ 
    Text and topic classification & \cite{olalekan2022_machine_learning_for_naija}, \cite{adelani2023_masakhanews}, \cite{ayogu2020exploring_naive_bayes}, \cite{ogueji2021_small-data}, \cite{hedderich2020_transfer_learning_distant} & Machine learning; multilingual language pre-training & MasakhaNEWS; Common
Crawl Corpus; Custom data \\ 
    Text-to-Speech & \cite{isewon_grapheme_based}, \cite{boco2022_end2end}, \cite{Akintola_hci}, \cite{ogunremi2024_iroyinspeech}, \cite{bibletts}, \cite{dagba2016_design_tts}, \cite{developing_yoruba_corpus}, \cite{antenatal}, \cite{Akinwonm2021_prosodic}, \cite{aoga2016integration_maryTTS}, \cite{abdulkareem2016_yorcall} & Rule-based & \`{I}r\`{o}y\`{i}nSpeech; Prosodic Read Speech Corpus; BibleTTS; Antenatal Orientation Speech Dataset; Yor\`{u}b\'{a} Speech Corpus; Custom data \\ 
    Tone identification and recognition & \cite{bengono2024improving_tone_recognition}, \cite{sosimi2019standard_svm}, \cite{iyanda2015_statistical_zipf} & Machine learning; neural networks & Yor\`{u}b\'{a} Speech Corpus; Custom data \\

    \label{tab5}
\end{longtable}

\noindent
\begin{longtable}{p{4cm}|p{1cm}p{2cm}p{2cm}p{1.8cm}p{2cm}p{2.6cm}|}
\caption{Cross Tabulation of NLP Tasks and Techniques} \\

\hline
\rowcolor[HTML]{87CEEB} 
\textbf{NLP Task} & 
\multicolumn{6}{c|}{\textbf{Techniques}}\\ 
\cline{2-7}

\rowcolor[HTML]{B0E0E6} 
 & Rule-Based & Statistical Modelling & Machine Learning & Deep Learning & Transfer Learning & Multilingual Pre-training \& Cross-Lingual Transfer\\ 
\hline
\endfirsthead

\hline
\rowcolor[HTML]{87CEEB}
\textbf{NLP Task} & 
\multicolumn{6}{c|}{\textbf{Techniques}} \\ 
\cline{2-7}

\rowcolor[HTML]{B0E0E6}
 & Rule-Based & Statistical Models & Machine Learning & Deep Learning & Transfer Learning & Multilingual Pre-training  \& Cross-Lingual Transfer \\ 
\hline
\endhead

\hline
\multicolumn{7}{r}{\textit{Continued on next page...}} \\ 
\hline
\endfoot

\hline
\endlastfoot

Acoustic unit discovery & \text{-} & \checkmark & \checkmark & \text{-} & \text{-} & \text{-}\\ \hline
    Automatic speech recognition & \text{-} & \checkmark &  \text{-} & \text{-}  & \checkmark & \text{-} \\ \hline
    Diacritic restoration & \text{-}& \checkmark &\checkmark & \checkmark & \text{-} & \text{-} \\ \hline
    Language modelling & \text{-} & \text{-} &  \text{-} & \checkmark  & \checkmark & \checkmark \\ \hline

    Information retrieval &  \text{-} & \text{-} & \text{-} & \text{-} & \checkmark & \checkmark  \\ \hline
    Machine translation    & \checkmark & \checkmark  &  \text{-} & \checkmark & \checkmark & \checkmark  \\ \hline

    Named Entity Recognition  & \text{-} & \text{-}  &  \checkmark & \text{-} & \checkmark & \checkmark  \\ \hline
    
    Part-of-speech tagging  & \text{-} & \checkmark  &  \checkmark & \text{-} & \checkmark & \checkmark  \\ \hline
    Sentiment analysis  & \checkmark & \checkmark  &  \checkmark & \checkmark & \checkmark & \checkmark  \\ \hline
    Speech-based gender recognition  & \text{-} & \text{-}  &  \text{-} & \checkmark & \text{-} & \text{-}  \\ \hline
    Spell checking \& correction  & \checkmark & \checkmark  &  \text{-} & \text{-} & \text{-} & \text{-}  \\ \hline
    Syllabification  & \checkmark & \text{-}  &  \checkmark & \checkmark & \text{-}  & \text{-}   \\ \hline
    Speech-to-Text  & \text{-}  & \checkmark  &  \text{-} & \text{-}  & \checkmark  & \text{-}  \\ \hline
    Text and topic classification  & \text{-} & \text{-}  &  \checkmark & \checkmark & \checkmark & \checkmark  \\ \hline
    Text-to-Speech  & \checkmark & \checkmark  &  \checkmark & \checkmark & \checkmark & \checkmark  \\ \hline
    Tone identification and recognition &  \text{-} &  \text{-} &  \checkmark & \checkmark &  \text{-} &  \text{-} \\
    \hline
\label{tab6}
\end{longtable}

\noindent
\textbf{AfriSenti}: AfriSenti \cite{muhammad2023_afrisenti} focuses on providing a high-quality, large-scale dataset to address the lack of resources for sentiment analysis in underrepresented African languages. This benchmark contains over $110,000$ posts involving $14$ African languages across four language families---Niger-Congo, Afro-Asiatic, English Creole and Indo-European---in which Yor\`{u}b\'{a} is a core part. The languages featured in the dataset are, namely:  Amharic, Algerian Arabic, Hausa, Igbo, Kinyarwanda, Moroccan Arabic, Mozambican Portuguese, Nigerian Pidgin, Oromo, Swahili, Tigrinya, Twi, Xitsonga, and Yor\`{u}b\'{a}. The sentiment categories captured in the corpus include positive, negative, and neutral. Furthermore, baseline traditional machine learning and pre-trained multilingual language models were trained to facilitate empirical analysis. Moreover, evaluation metrics include accuracy, F1 score, and precision/recall. Overall, it established a benchmark for sentiment analysis in Yor\`{u}b\'{a} language, also enabling comparisons of models and techniques for African languages.\\

\noindent
\textbf{AfriWOZ}: AfriWOZ \cite{adewumi2023afriwoz} also pioneers in resource development involving dialogue generation for low-resource African languages. It presents a high-quality dialogue generation dataset for $6$ African languages, emphasising the Yor\`{u}b\'{a} language as a subset. The other five languages include Hausa, Kinyarwanda, Nigerian Pidgin English, Swahili, and Wolof; all languages belong to three language families: Afro-Asiatic, Niger-Congo, and English Creole. The dialogue dataset for the included languages was developed from the MultiWoz dataset \cite{budzianowski-etal-2018-multiwoz}, and experiments were performed to provide empirical analysis for related tasks using transfer learning approaches. \\ 

\noindent
\textbf{Antenatal Orientation Speech Dataset}: This dataset comprises multilingual speech samples in English, Nigerian Pidgin English, and Yor\`{u}b\'{a} languages, primarily for antenatal health education in Nigeria, which is the primary geographical source of these languages.  The dataset \cite{antenatal} is composed of a word count of $2,639$ for English, $3,202$ for Yor\`{u}b\'{a}, and $2,521$ for Pidgin. Moreover, the corresponding speech datasets had $59,880$ seconds for English, $70,380$ seconds for Yor\`{u}b\'{a}, and $69,840$ seconds for Pidgin, both showing a dominant percentage for the Yor\`{u}b\'{a} language. The size of the speech datasets was $15.6$ KB for English, $17.9$ KB for Yor\`{u}b\'{a}, and $18.3$ KB for Pidgin. This dataset will be valuable for antenatal orientation in Nigeria and contribute to the currently low-resource status of  Yor\`{u}b\'{a} language by improving its availability for NLP tasks.

\noindent
\textbf{BibleTTS}: BibleTTS \cite{bibletts} is an open speech multilingual dataset containing ten languages of sub-Saharan Africa, viz: Akuapem Twi, Asante Twi, Chichewa, Ewe, Hausa, Kikuyu, Lingala, Luganda, Luo, and Yor\`{u}b\'{a}. The corpus emanates from Bible recordings provided by the Open-Bible project of Biblica and contains high-quality studio recordings for single speakers, with up to $86$ hours of data per language. Furthermore, TTS models were developed by leveraging the dataset, which ascertains their usability and quality for speech synthesis. The dataset is openly available for commercial and non-commercial use as it is licensed under the CC-BY-SA License, promoting accessibility and further research in text-to-speech for Yor\`{u}b\'{a}, among other languages captured. \\

\noindent
\textbf{CIRAL}: CIRAL\cite{adeyemi2024_ciral} presents a cross-lingual information retrieval (CLIR) in four languages---Hausa, Somali, Swahili, and Yor\`{u}b\'{a}---with great emphasis on Yor\`{u}b\'{a} language. The corpus comprises over $2.5$ million in passages curated from indigenous African websites.  The data annotation involved native speakers annotating over $1,600$ queries and $30,000$ binary relevance judgments, ensuring high-quality data for evaluation. Furthermore, reproducible baselines involving translation-based and embedding-based CLIR were further developed to support empirical analysis for related tasks. The dataset is relevant for retrieving relevant documents or passages in one language, e.g., Hausa, Swahili, Yor\`{u}b\'{a}, or Somali, based on queries expressed in another language. \\

\noindent
\textbf{IkiniYor\`{u}b\'{a}}: IkiniYor\`{u}b\'{a}\cite{akinade2023_varepsilon} is a dataset focusing on the cultural aspect of Yor\`{u}b\'{a}, intending to investigate how well MT models can translate greetings in the language into English. This dataset comprises $960$ parallel sentences of Yor\`{u}b\'{a} greetings and their English translations. Moreover, it also incorporates the movie transcript subset of the MENYO-20k dataset\cite{adelani2021effect_domain_and_diacritic}, which contains conversational Yor\`{u}b\'{a}-English data. The performance of MT models was evaluated on the datasets to enable empirical analysis for related studies in the future. The introduction of the IkiniYor\`{u}b\'{a} dataset is a valuable contribution to the research community working on Yor\`{u}b\'{a} NLP.\\

\noindent
\textbf{\`{I}r\`{o}y\`{i}nSpeech}: \`{I}r\`{o}y\`{i}nSpeech \cite{ogunremi2024_iroyinspeech} presents high-quality speech corpus exclusively in Yor\`{u}b\'{a} language, which allows for multipurpose usage involving ASR and TTS tasks. The data collection involves curating 23,000 sentences from news and creative writing domains under an open license (CC-BY-4.0). Also, it included 5,000 sentences on the Mozilla Common Voice platform to crowdsource recordings and validations. The data contributions include 42 hours of in-house recorded speech data by 80 volunteers and an additional 6 hours of validated recordings on Mozilla Common Voice. Ultimately, a high-fidelity single-speaker Yor\`{u}b\'{a} TTS system was evaluated on the curated speech data, including a baseline word error rate (WER) of $23.8$ for ASR purposes. \\

\noindent
\textbf{Lagos-NWU Yor\`{u}b\'{a} Speech Corpus}: To promote research voice recognition, a speech corpus including $16$ female and $17$ male speakers was recorded in Lagos, Nigeria \cite{enikuomehin_implementation_lagos_women}. A total of $130$ utterances read from brief texts chosen for phonetic coverage were recorded by each speaker. Moreover,  recordings were made using a microphone attached to a laptop computer in a peaceful office setting to ensure quality in the corpus. Ultimately, this corpus will be relevant for various NLP tasks in Yor\`{u}b\'{a}, such as TTS, ASR, etc. \\

\noindent
\textbf{MasakhaNER}: MaskhaNER \cite{adelani2021_masakhaner_1} boasts as the first largest high-quality dataset for named entity recognition tasks focused on African languages, in which Yor\`{u}b\'{a} was greatly emphasised.  The dataset scope extends to $10$ African languages and across four language families, and they include Yor\`{u}b\'{a}, Amharic, Hausa, Igbo, Kinyarwanda, Wolof, Nigerian-Pidgin, Luo, and Swahili. \\

\noindent
\textbf{MasakhaPOS}: MasakhaPOS \cite{dione2023_masakha_pos} presents a huge POS dataset available for $20$ diverse African languages, of which Yor\`{u}b\'{a} language was greatly emphasised. Other languages featured in the dataset include Bambara, Fon, Hausa, Igbo, Luo, Luganda, Akan/Twi, Wolof, isiZulu, isiXhosa, Setswana, Kiswahili, Nigerian Pidgin English, Mossi, Kinyarwanda, Ghom\'{a}l\'{a}, \'{E}w\'{e}, chiShona, and Chichewa.  Moreover, Yor\`{u}b\'{a} data was jointly obtained from Voice of Nigeria and Asejere online newspapers, containing a $43,601$ token with an average sentence length of $24.4$. Baseline models using conditional random fields (CRF) and multilingual PLMs such as mBERT and XLM-R were developed to support empirical analysis for related tasks.\\

\noindent
\textbf{MENYO-20k}: MENYO-20k \cite{adelani2021effect_domain_and_diacritic} presents a multi-domain parallel corpus of Yor\`{u}b\'{a}-English with clean orthography and standardized train-test splits, with focus on improving the evaluations of MT models on low-resource languages. The dataset's domain includes texts from news articles,  radio and movie transcripts, and TED talks.  In addition,  special attention was given to the diacritization of Yor\`{u}b\'{a} texts in the corpus, as it plays an essential role in intelligibility and translation quality in MT NLP tasks. Neural MT models are benchmarked to enable future empirical analysis in similar studies. The dataset is available through a CC BY-NC 4.0 licence.\\

\noindent
\textbf{NaijaSenti}: NaijaSenti \cite{muhammad2022_naijasenti} is a sentiment classification corpus for four prominent languages used as a medium of communication in Nigeria: Yor\`{u}b\'{a}, Hausa, Nigerian Pidgin English and Igbo. The dataset was obtained from X---formerly Twitter--- and comprises roughly $30,000$ tweets for each language, including code-mixed tweets. Benchmarks were also developed on monolingual sentiment analysis tasks for each language, and the dataset and model codes are publicly available online. \\

\noindent
\textbf{NollySenti}: NollySenti \cite{shode2023_nollysenti} is a parallel multilingual corpus for sentiment classification in five languages spoken in Nigeria, including Yor\`{u}b\'{a}, Nigerian Pidgin English, Hausa, Igbo, and English. The dataset source reviews from Nollywood movies---movies primarily made in Nigeria---initially in English, with translations into the four Nigerian languages involved. Moreover, the initial collection for the dataset comprises $882$ negative reviews and $1,018$ positive reviews, with an average word length of $20.7$ and $35.0$, respectively, making up two sentiment classification categories.  Benchmarks were also developed through traditional machine-learning techniques and pre-trained language modelling to enable empirical analysis for future research in the domain.  \\

\noindent
\textbf{Prosodic Read Speech Corpus}: This corpus contains a high-quality speech corpus \cite{Akinwonm2021_prosodic} of $7,376$ phrases and sentences in Standard Yor\`{u}b\'{a} language. A TTS system was developed on the corpus and evaluated using the mean open score and the semantically unpredictable sentence (SUS) test score to support empirical analysis for future related studies. The developed corpus can be used as valuable research material for future studies on Yor\`{u}b\'{a} speech processing, synthesis, and recognition\\

\noindent
\textbf{TATA}: TATA \cite{gehrmann2023_tata} dataset is a large and high-quality multilingual dataset focused on African languages, created from Demographic and Health Surveys (DHS) Program bilingual reports. It consists of $8,700$ examples in nine languages, with Russian as a zero-shot test language.  Yor\`{u}b\'{a} is a core component of these languages; other languages in the dataset include Arabic, English, French, Hausa, Igbo, Portuguese, and Swahili. A transformer-based multilingual PLM was evaluated on the dataset, supporting empirical analysis in future related studies.\\

\noindent
\textbf{TTS Yor\`{u}b\'{a} Speech Corpus}: A comprehensive Yor\`{u}b\'{a} speech corpus was designed primarily for TTS synthesis research and development in the language \cite{dagba2016_design_tts}. The Yor\`{u}b\'{a} speech corpus contains $2,415$ sentences with $46,117$ words and $148,823$ phonemes. The corpus has a good balance of sentence types---affirmative, interrogative, and exclamatory---and phoneme distribution. Furthermore, the Yor\`{u}b\'{a} speech corpus was also integrated into the MaryTTS\footnote{\url{https://marytts.github.io/}} open-source multilingual text-to-speech (TTS) synthesis platform, which achieved a Mean Opinion Score (MOS) of $2.9$ out of $5$ for the quality of the synthesized speech.\\

\noindent
\textbf{WebCrawl African}: WebCrawl African \cite{vegi2022_webcrawl_african} contains multilingual parallel corpora for diverse African languages. It was gathered to build resources for machine translation tasks in low-resource and extremely low-resource African languages. The parallel corpus spans 74 language pairs, which includes $15$ African languages, of which Yor\`{u}b\'{a} language was greatly emphasised. For empirical analysis purposes, two multilingual models were trained on behalf of $24$ African languages, including Yor\`{u}b\'{a} language,  using the dataset and evaluated using FLORES-200 \cite{nllbteam2022languageleftbehindscaling_flores200} benchmarks. The dataset will be useful for multilingual and cross-lingual machine translation tasks involving Yor\`{u}b\'{a} language. \\

\noindent
\textbf{W\'{U}R\`{A}}: W\'{U}R\`{A} \cite{oladipo2023_better} presents a new high-quality multilingual pre-training corpus for African languages, with Yor\`{u}b\'{a} greatly emphasised. The dataset is $\approx 30$GB for all languages and  $\approx 19$GB for African languages. Downstream tasks ranging from MT, summarization, cross-lingual question answering, and text classification were built on the corpora to enable empirical analysis for related future studies. \\

\noindent
\textbf{YFACC}: YFACC \cite{olaleye2023_yfacc} connected speech with visual representations in the dataset comprising spoken captions in Yor\`{u}b\'{a} language for $6,000$ Flickr\footnote{\url{https://www.flickr.com}} images. The audio captions also utilised single-speaker recording to ensure consistency in the audio data. Moreover, it includes cross-lingual pairing, whereby images are tagged with English visual labels and paired with Yor\`{u}b\'{a} speech, permitting cross-lingual applications. The empirical analysis is also catered for by developing a baseline cross-lingual model. Ultimately, this dataset addressed the dearth of visually-grounded speech datasets and a new benchmark in low-resource languages, specifically the Yor\`{u}b\'{a} language. \\

\noindent
\textbf{Yor\`{u}b\'{a} Speech Corpus}: This contains an open-source speech dataset exclusively for Yor\`{u}b\'{a} \cite{developing_yoruba_corpus}. This corpus comprises over $4$ hours of recordings from $36$ male and female volunteers. It also includes transcriptions with disfluency annotations and full diacritization, essential for pronunciation and lexical disambiguation. Moreover, the dataset was tested in a statistical parametric speech synthesis (SPSS) for evaluation purposes and compared with related corpora in the same domain.  The corpus supports TTS, ASR, and speech-to-speech translation, contributing to West African corpus linguistics.\\

\noindent 
\textbf{Yor\`{u}b\'{a} Treebank (YTB)}: YTB \cite{ishola2020_yoruba_dependency_treebank} boasts of the first-tree bank in Yor\`{u}b\'{a} language. It contains manually annotated Yor\`{u}b\'{a} Bible sections and is relevant for investigating dependency analysis in the language. \\

\noindent
\textbf{YOR\`{U}LECT}: YOR\`{U}LECT \cite{ahia2024voices_unheard} introduces a high-quality parallel text and speech corpus specifically in Yor\`{u}b\'{a} language, across three domains---machine translation, automatic speech recognition, and speech-to-text translation---and four Yor\`{u}b\'{a} dialects---Standard Yor\`{u}b\'{a}, If\d{\`{e}}, \`{I}j\d{\`{e}}b\'{u}, and \`{I}l\`{a}j\d{e}. The If\d{\`{e}} dialect is spoken primarily among the people of \d{\`{O}}\d{s}un state, \`{I}j\d{\`{e}}b\'{u} among people of \d{\`{O}}g\`{u}n state, and \`{I}l\`{a}j\d{e}, among the people of O\`{n}d\'o state; all states located in the South West geopolitical zone of Nigeria. Text data from various sources were obtained and localised into the three Yor\`{u}b\'{a} dialects for the corpus development. Similarly, high-quality speech data in standard  Yor\`{u}b\'{a}, If\d{\`{e}}, and \`{I}l\`{a}j\d{e} were recorded.  The text portion contains about $1506$ parallel sentences for each dialect, totalling $6024$, and the speech part contains $\approx3$ hours of audio for the considered dialects. Furthermore, to aid empirical analysis, experiments involving the domains of MT, ASR, and speech-to-text were done. The dataset and models were published under an open licence, making it relevant for Yor\`{u}b\'{a} NLP tasks.

\noindent
Table \ref{tab7} is used to summarize the available language resources relevant for NLP development involving the Yor\`{u}b\'{a} language.

\begin{longtable}{p{3cm}p{4cm}p{4cm}p{2cm}p{2cm}}
\caption{Summary of Resources Available for Yor\`{u}b\'{a} NLP Development} \\ 

    \hline
    \rowcolor[HTML]{87CEEB} 
    Resource & Type & Use Case & Language Frequency & Study Used \\ \hline
    \endfirsthead
    \hline
    \rowcolor[HTML]{87CEEB} 
    Resource & Type & Use Case & Language Frequency & Study Used \\ \hline
    \endhead
    \hline
    \multicolumn{5}{|r|}{\textit{Continued on next page...}} \\ \hline
    \endfoot
    \endlastfoot

    AFRIQA    & Annotated text corpus  & Cross-lingual open-retrieval question answering & Multilingual & \cite{ogundepo-etal-2023-cross} \\
    AfriSenti & Annotated text corpus & Sentiment analysis & Multilingual & \cite{muhammad2023_afrisenti} \\ 
    AfriWOZ & Dialogue dataset & Dialogue generation & Multilingual & \cite{adewumi2023afriwoz} \\     Antenatal orientation & Speech corpus, text corpus & Speech recognition, machine translation & Multilingual & \cite{antenatal} \\
    BibleTTS & Speech corpus & Text-to-speech & Multilingual & \cite{bibletts} \\ 
    CIRAL & Annotated text corpus & Cross-lingual information retrieval & Multilingual & \cite{adeyemi2024_ciral} \\ 
    IkiniYor\`{u}b\'{a} & Parallel corpus & Machine translation  & Bilingual & \cite{akinade2023_varepsilon} \\ 
    \`{I}r\`{o}y\`{i}nSpeech & Speech corpus & Automatic speech recognition, text-to-speech & Monolingual & \cite{ogunremi2024_iroyinspeech} \\ 
    Lagos NWU Yor\`{u}b\'{a} Speech Corpus & Speech corpus & Automatic speech recognition & Monolingual & \cite{enikuomehin_implementation_lagos_women}, \cite{alqahtani2019investigating_input-output_diacritic}, \cite{alqahtani2019_efficient_CNN}, \cite{orife_sequence2sequence} \\ 
    MasakhaNER & Annotated text corpus & Named entity recognition & Multilingual & \cite{adelani2021_masakhaner_1} \\ 
    MasakhaPOS & POS dataset & Part-of-speech tagging & Multilingual & \cite{dione2023_masakha_pos} \\ 
    MIRACL & Annotated text corpus & Monolingual information retrieval & Multilingual & \cite{zhang2023_miracl}, \cite{oladipo2024_backbones} \\
    MENYO-20k & Parallel corpus & Machine translation & Bilingual & \cite{adelani2021effect_domain_and_diacritic}, \cite{adeboje_bilingual_transformer}, \cite{signoroni2023_evaluating_sentence_alignment} \\ 
    NaijaSenti & Annotated text corpus & Sentiment analysis & Multilingual & \cite{muhammad2022_naijasenti} \\ 
    NollySenti & Parallel corpus & Sentiment classification, machine translation & Multilingual & \cite{shode2023_nollysenti} \\
    Prosodic Read Speech & Speech corpus & Text-to-speech & Monolingual & \cite{Akinwonm2021_prosodic} \\ 
    TATA & Parallel corpus & Data-to-text generation, multilingual generation & Multilingual & \cite{gehrmann2023_tata} \\ 
    TTS Yor\`{u}b\'{a} Speech Corpus & Speech corpus & Text-to-speech  & Monolingual & \cite{dagba2016_design_tts} \\ 
    Webcrawl African & Parallel corpora & Machine translation & Multilingual & \cite{vegi2022_webcrawl_african} \\
    W\'{U}R\`{A} & Parallel corpus & Cross-lingual question answering; text classification; summarization; \& machine translation & Multilingual & \cite{oladipo2023_better} \\
    YFACC & Speech-image dataset & Cross-lingual keyword localisation & Monolingual & \cite{olaleye2023_yfacc} \\
    Yor\`{u}b\'{a} Speech Corpus & Speech corpus & Text-to-speech; automatic speech recognition; speech-to-speech translation & Monolingual & \cite{developing_yoruba_corpus, bengono2024improving_tone_recognition}, \cite{ondel2022non_bayesian}, \cite{yusuf_hierarchical} \\ 
    
    Yor\`{u}b\'{a} Treebank & Dependency treebank & Dependency parsing and analysis & Monolingual & \cite{ishola2020_yoruba_dependency_treebank} \\ 

    YOR\`{U}LECT & Parallel text and speech corpus & Machine translation; automatic speech recognition; \& speech-to-text translation & Monolingual & \cite{ahia2024voices_unheard}\\
    
    \hline
    \label{tab7}
\end{longtable}

\noindent
\textbf{Synthesis and Implications of Available Resources} \\

While the listed resources reflect commendable progress in Yor\`{u}b\'{a} NLP, there exist significant limitations that shape the scope of current research. For instance, although multiple corpora exist, most are either small-scale, domain-specific, or lack detailed linguistic annotation, which hinders their use for advanced syntactic or semantic modeling tasks. The absence of standard benchmarks further complicates reproducibility and model comparison.

\noindent
Moreover, lexical and morphological resources such as POS taggers primarily support rule-based or traditional machine learning methods. In contrast, resources suited for training modern transformer-based models---such as large-scale annotated datasets or pretrained embeddings specifically for Yor\`{u}b\'{a}---are sparse or underutilized. In addition, limited speech resources restrict progress in speech-to-text or conversational systems.

\noindent
These gaps suggest that while foundational resources exist, their fragmented and often narrow nature limits scalability and generalization. Going forward, resource creation should further prioritize open licensing, broader domain coverage, and alignment with task-specific benchmarks to better support the evolving landscape of low-resource NLP.

\subsubsection{RQ4: What are the major challenges in developing NLP solutions for Yor\`{u}b\'{a}? }
The development of natural language solutions for Yor\`{u}b\'{a} faces multiple challenges, which hinder the language's representation and accessibility in digital and computational contexts, especially as a low-resource language. Data synthesis in this study has helped identify five primary challenges, including linguistic, technical, resource, cultural, and societal factors, and evaluation and benchmarking challenges. \\

\noindent
\textbf{Linguistic challenges}: The Yor\`{u}b\'{a} language exhibits several linguistic properties that make NLP development in it an exigent task.  These include its tone-dependence, complex morphology, diacritics dependency, and code-switching language usage. Each of these challenges is briefly described below:
\begin{itemize}
    \item \textbf{Tonal complexity}: Yor\`{u}b\'{a} language's implicit pitch contours, denoting tonal components, is used in communicating emphasis and other para-linguistic expressions. Thus, they usually determine the semantic property of the message conveyed in the language. For instance, \textit{\`{O}g\'{u}n} (a deity associated with iron) is different from either of \textit{ogun} (war) and \textit{og\'{u}n} (twenty). This distinction in linguistic meaning complicates various NLP tasks even though the words belong to the same phonetic sequence. For instance, one of the main challenges for speech recognition in Yor\`{u}b\'{a}  is determining the tone associated with a syllable \cite{bengono2024improving_tone_recognition}. Similarly, utilising context-dependent phone units to capture long-term spectrum dependencies of tone in Yor\`{u}b\'{a} is typically less successful and oftentimes requires a different means of acoustic modelling \cite{sosimi2019standard_svm}. Furthermore, in bilingual English-Yor\`{u}b\'{a} MT tasks, tone-changing verbs \cite{I__2015_tone_changing} also present challenges as they frequently alter the semantic properties of the sentence they are used in by shifting from a low-tone to a mid-tone. Consequently, accurate representation and processing of tones are essential for NLP tasks like diacritic restoration, TTS synthesis, and MT.
    \item \textbf{Diacritic dependency}: Similar to the tonal feature of Yor\`{u}b\'{a} language, it also requires appropriate assignment of accent property to characters in a segment. This task is usually referred to as diacritic restoration, and it is essential to fully decipher the linguistic meaning of words. Diacritic marks are usually placed above, below, or between characters to indicate pronunciation and may change the meaning of the composed words \cite{alqahtani2019investigating_input-output_diacritic}.  Like most languages involving diacritics, Yor\`{u}b\'{a} language users often omit them in electronic texts, increasing lexical ambiguity and also posing challenges to NLP  systems as a result of mislaying the accompanying syntactic, grammatical, or
semantic information, partially or totally \cite{ayogu2021_diacritic}. This omission is sometimes due to the unavailability of supporting applications and devices \cite{orife_sequence2sequence} or lack of knowledge by most users \cite{abdulkareem2016_yorcall}, resulting in a drawback towards improving NLP representation in the language. 
\item \textbf{Morphological Complexity}: Carrying out analysis of word formation and structure in Yor\`{u}b\'{a} language tends to be oftentimes complex due to its plenitude in terms of rules involving noun and verb inflection patterns \cite{agbeyangi2016_morphological_standard_yoruba_nouns}. Handling agglutinative morphology and word compounding has also been stated to require advanced morphological analysis tools \cite{adegbola2016_pattern-based}. The deficiency in this regard frequently creates ambiguity in unsophisticated NLP systems due to issues like affixes of words and the required rules to correctly program the system for effective analysis. Efforts towards addressing this have incorporated automatic morphological induction \cite{adegbola2022_automatic_morphology} to ensure compatibility, producing a more accurate representation in the NLP domain. 

\item \textbf{Code-Switching}: This is common for most Nigerian languages, and Yor\`{u}b\'{a} is not spared of the growing `civilization'. Despite initiatives towards sustaining native languages, it is evident that the younger generation finds it difficult to carry on lengthy conversations or even to form lengthy sentences without interspersing the conversation with terms frequently borrowed from the English language \cite{adewole2020_vowel_elision}. This phenomenon introduces additional complexity in NLP tasks like language identification, sentiment analysis, and machine translation, as it reduces data availability purely in the language. This is evident in the dataset obtained for NaijaSenti \cite{muhammad2022_naijasenti}, which aimed to conduct monolingual sentiment analysis for four languages through a curated corpus from X, as a certain proportion of the corpus had to cater to the phenomenon. 

\end{itemize}

\noindent
\textbf{Resource challenges}: The challenges in obtaining substantial or quality resources for training NLP models involving Yor\`{u}b\'{a} language are the ultimate pitfall for their NLP development, just like in many other low-resource languages \cite{oladipo2023_better, adelani2021effect_domain_and_diacritic}. The existence of a comprehensive lexicon for most high-resource languages makes their pre-processing task less taxing, unlike a low-resource language like Yor\`{u}b\'{a}. The unavailability often leads to applying manual data cleaning, hence the task of token validation for corpus development \cite{adewole2017_token_validation}. \\

\noindent
In addition, limited corpora pose significant challenges in various NLP tasks.  For instance, a comparatively short corpus of speech recordings and minimal language-specific development can automatically build acoustic models for an elementary speech synthesiser in a new language; however, tonal languages like  Yor\`{u}b\'{a} require additional resources \cite{van2014predicting_utterance_pitch_target}, which is usually not readily available. Also, word-level and character-level models are especially common for most diacritic restoration tasks \cite{alqahtani2019investigating_input-output_diacritic}; however, they require significant training data to prevent sparsity in the models. Similarly, out-of-vocabulary word translation is a significant issue for low-resource languages that lack parallel training data \cite{liu2018_context_OOV}. This phenomenon is considered a stumbling block to NLP development in these different domains, albeit recent efforts have been directed towards implementing other viable methods requiring lower resources for a handful of tasks supporting such. \\

\noindent
Generally, limited available corpora for developing NLP tools have been a consistent challenge in NLP development involving Yor\`{u}b\'{a} and under low-resource languages. Studies have also shown the presence of quality issues for existing corpora and models \cite{oladipo2023_better}, which limits the reliability of findings from using such corpora.  Consequently, efforts towards creating readily available corpora for specific NLP tasks are sacrosanct since better results mostly require high-quality and large data \cite{ayogu2021_diacritic}. \\

\noindent
\textbf{Technological challenges}: 
Limited availability of pre-processing tools is also a phenomenon constituting a major setback for most low-resource African languages, as the tool suitable for a language tends not to suit another, such as Yor\`{u}b\'{a}, owing to its morphological complexity and diacritic dependency. For instance, English word punctuation segmentation will not necessarily work for \textit{Arabizi}, the Arabic chat language, since these marks define its own orthography \cite{AlBadrashiny2016_SPLITS_preprocessing}.  Furthermore, research on pre-processing tools for different languages within the same family collectively suggests that while some can work across related languages, optimal results often require language-specific adjustments and careful consideration of the target domain and linguistic features \cite{uysal2014_text_preprocessing}.  \\

\noindent
Other language tools, including the syntactic parser and POS tagger, have also been reported to mostly fail in accounting for Yor\`{u}b\'{a} language linguistic nuances \cite{Abiola_hidden_markov}, thereby constituting a challenge requiring the development of language-specific tools for these specific tasks. Similarly, an essential NLP task such as effective spell-checking development in Yor\`{u}b\'{a} is still at the early development \cite{asahiah2023_diacritic} due to the non-consideration of diacritic necessity in earlier studies \cite{oladiipo2020_spelling_error_patterns}, resulting in their limitation.  A limited number of studies \cite{dione2023_masakha_pos, oluwaseyi2024_spelling, asahiah2023_diacritic} involving computational NLP have been seen to include Yor\`{u}b\'{a} language in these domains. Furthermore, the majority of Yor\`{u}b\'{a} texts are also seen to be typed using plain ASCII, without diacritics \cite{orife_sequence2sequence}, perhaps due to limited tools supporting their full implementation. This situation also induces lower quality translation between Yor\`{u}b\'{a} language and most European languages due to a lack of an adequate elision resolution tool \cite{adewole2020_vowel_elision}. Thus, there is a growing need for flexible multilingual pre-processing solutions.  \\

\noindent
\textbf{Cultural and Societal Factors}: An African language, such as Yor\`{u}b\'{a}, is not only a mirror into the mind of its users but also a mirror into their culture and history \cite{okediya2019_building_ontology}. This emphasises the richness of the culture and language. However, younger generations of Yor\`{u}b\'{a} speakers increasingly favour English for formal and digital communication, while the parents also decide not to teach their infants many times \cite{abdulkareem2016_yorcall}. This shift reduces the volume of contemporary Yor\`{u}b\'{a} texts and contributes to the decline of the language in digital contexts. Moreover, despite the evident progress in MT tasks for low-resource languages, NMT
models still lag in accurately carrying out automatic translation involving cultural references \cite{adelani2021effect_domain_and_diacritic} for Yor\`{u}b\'{a} language. Consequently, cultural and societal challenges of Yor\`{u}b\'{a} in NLP development require considerable attention, as they could potentially endanger the language in the face of westernization, globalization, and inter-ethnic communication \cite{okediya2019_building_ontology}.\\

\noindent
Table \ref{tab8} summarizes the challenges of NLP development involving Yor\`{u}b\'{a} language and some of the highlighted solutions.

\begin{table*}[htbp]
    \centering
    \setlength{\arrayrulewidth}{0.1mm}
    \setlength{\tabcolsep}{8pt}
    \renewcommand{\arraystretch}{1.5}
    \arrayrulecolor[HTML]{000000} 
    \caption{Summary of Challenges in Yor\`{u}b\'{a} NLP Development}
    \begin{tabular}{p{4cm}p{4cm}p{4cm}p{4cm}}
        \hline
        \rowcolor[HTML]{87CEEB} 
        Challenge category & Specific challenge & Primary studies & Proposed solutions \\
        \hline
    
        Linguistic    & Tonal complexity, diacritic dependency, morphological complexity, \& code-switching & \cite{I__2015_tone_changing}, \cite{bengono2024improving_tone_recognition}, \cite{van2014predicting_utterance_pitch_target}, \cite{sosimi2019standard_svm}, \cite{orife_sequence2sequence}, \cite{ayogu2021_diacritic}, \cite{alqahtani2019_efficient_CNN}, \cite{alqahtani2019investigating_input-output_diacritic}  & Diacritic-aware models;
        Automatic morphological induction \\
        Technological & Limited pre-trained models and lack of language-specific tools   & \cite{ogueji2021_small-data}, \cite{fagbolu2016_applying_rough_set}, \cite{adewole2020_vowel_elision} & Fine-tuning multilingual models; Developing elision resolution tool \\ 
        Resource scarcity & Limited annotated corpora for specific tasks    & \cite{ayogu-ner},\cite{oladipo2023_better}, \cite{adewumi2023afriwoz}, \cite{adewole2020_vowel_elision}  & Collaborative corpus development across various NLP tasks;
        Multilingual corpus development

        \\ 
        Socio-cultural & Adopting foreign language as first language  &  \cite{abdulkareem2016_yorcall}, \cite{okediya2019_building_ontology} & Implementing iterative learning systems for  Yor\`{u}b\'{a} language \\ \hline
    \end{tabular}
    \label{tab8}
\end{table*}

\section{Discussion}
\subsection{General Overview}
This systematic review was intended to capture the growth and current stage of NLP participation for Yor\`{u}b\'{a} language over a decade. Through the information synthesised from the primary studies, it is obvious that significant progress has been made in notable areas of NLP involving Yor\`{u}b\'{a} language.  MT has particularly received more attention through the inclusion of the language primarily in bilingual MT research \cite{safiriyu2015_personal_pronouns, babatunde2024_speech-to-text_hybrid, signoroni2023_evaluating_sentence_alignment, Esan_2020_rnn, adeboje_bilingual_transformer, babatunde2021english_android_phones, elesemoyo_numeral} involving translation between two languages. Related studies investigating the inherent prerequisite towards enhancing translation qualities in MT model qualities, the development of MT tools, such as vowel elision resolution \cite{adewole2020_vowel_elision}, and the application of rough set theory \cite{fagbolu2016_applying_rough_set} in translation, have also been explored. Moreover, the need to create better quality data \cite{oladipo2023_better} has also led towards improved research in this domain through the development of additional parallel corpora and leveraging multilingual language pre-training to improve the model training capability in the presence of large high-quality datasets. \\

\noindent
Similar to MT research, significant efforts have been made in TTS synthesis, with considerable efforts towards corpus development, as evidenced by publicly available datasets, most of which are speech corpora. Even though it is the primary focus in most research investigating the task, some of the studies have included ASR and speech-to-text \cite{ogunremi2024_iroyinspeech, Akintola_hci, ahia2024voices_unheard} in their experiments, contributing to the entire research. Furthermore, sentiment analysis and information retrieval tasks involving Yor\`{u}b\'{a} have also benefited significantly from multilingual corpus development through the need to develop language resources for low-resource African languages. Most have been through the works in \cite{muhammad2023_afrisenti} and \cite{adeyemi2024_ciral}. \\

\noindent
Generally, NLP tasks such as sentiment analysis, machine translation, text-to-speech synthesis, automatic speech recognition, named entity recognition, information retrieval, parts-of-speech tagging, and language modelling have received considerable research efforts, with at least four studies each addressing them. In addition, innovative approaches, transfer learning, and diacritic-aware systems have demonstrated promising results, showcasing the adaptability of state-of-the-art techniques to Yor\`{u}b\'{a} NLP, albeit earlier studies have depended mostly on the rule-based methods. Resources like Yor\`{u}b\'{a} speech corpus \cite{developing_yoruba_corpus}, MENYO-20k \cite{adelani2021effect_domain_and_diacritic}, Lagos NWU Yor\`{u}b\'{a} Speech Corpus \cite{enikuomehin_implementation_lagos_women} among others, have also been pivotal in advancing the field by providing foundational datasets, as they have been used in more than one study.\\

\noindent
Furthermore, recent studies have showcased the importance of multilingual and cross-lingual techniques, as they help to promote language availability for many African languages simultaneously. This is evident from named entity recognition \cite{adelani2021_masakhaner_1, adelani2022_masakhaner_2}, sentiment analysis \cite{muhammad2022_naijasenti, muhammad2023_afrisenti, shode2023_nollysenti}, and information retrieval tasks \cite{ogundepo-etal-2023-cross, adeyemi2024_ciral, zhang2023_miracl}. Also, models pre-trained on multilingual datasets have been seen to be beneficial for a low-resource language like Yor\`{u}b\'{a}. Overall, it is also evident that collaborative, open-source and community initiatives have greatly improved research and availability of resources in the bid to overcome resource scarcity whilst fostering knowledge-sharing among these resource-scarce African languages. \\

\noindent
However, in the context of constraints stunting the rapid growth of NLP in Yor\`{u}b\'{a} language, studies have highlighted mainly linguistic complexity such as diacritic dependency \cite{orife_sequence2sequence} and tonal variation \cite{I__2015_tone_changing}. Other challenges involving cultural and societal factors were also highlighted, including the primary hindrance, which has always been a limitation in the availability of corpora or the quality of such available corpora. \\

\noindent
Ultimately, the geographical distribution of lead authors for the studies shows promising insight into the research efforts towards promoting Yor\`{u}b\'{a} in NLP involvement is global and not restricted to the primary home of the language. It also revealed necessary collaborations among authors and its pivot towards collectively improving resource availability for African languages. \\

\noindent
This synthesis of findings not only elucidates the current state of Yor\`{u}b\'{a} NLP but also offers broader lessons for the development of NLP in other low-resource contexts. In many respects, the linguistic and infrastructural challenges identified here are mirrored in numerous other low-resource languages (LRLs). This makes it possible to extend the methodological insights gained from Yor\`{u}b\'{a} NLP to a wider set of languages with similar constraints.

\subsection{Implications for Other Low-Resource Languages}

Although this review focuses on the Yor\`{u}b\'{a} language, many of the observed trends, challenges, and methodological solutions are highly transferable to other LRLs, particularly those with similar linguistic complexity, socio-cultural context, or infrastructural limitations. Challenges such as dearth of annotated corpora, limited pre-trained language models, and tonal complexity are not unique to Yor\`{u}b\'{a}; they also hinder progress in languages like Igbo, Hausa, Wolof, Amharic, and Somali \cite{orife_sequence2sequence, mehari2024semi-supervised_NER}. \\

\noindent
From task NLP tasks perspective, diacritic restoration, which was addressed for Yor\`{u}b\'{a} language through neural network–based approaches \cite{orife_sequence2sequence, ayogu2021_diacritic, alqahtani2019_efficient_CNN} using datasets like the Lagos-NWU Yor\`{u}b\'{a} Speech Corpus could be adapted to other tone-marked or diacritic-dependent languages such as Ewe, Twi, or Vietnamese, as seen in the publications. Similarly, machine translation pipelines that combined rule-based, statistical, and neural methods \cite{adebara2022_linguistically_motivated, eludiora2016_eng2yor_MT, adeboje_bilingual_transformer, I__2015_tone_changing, babatunde2024_speech-to-text_hybrid, Esan_2020_rnn, liu2018_context_OOV, adebara-etal-2024-cheetah} and resources like MENYO-20k \cite{adelani2021effect_domain_and_diacritic} and IkiniYor\`{u}b\'{a} \cite{akinade2023_varepsilon} could be replicated for other LRLs with parallel corpora, enabling both bilingual and multilingual translation systems. Furthermore, in sentiment analysis tasks, multilingual resources such as AfriSenti \cite{muhammad2023_afrisenti} and NaijaSenti \cite{muhammad2022_naijasenti} have shown that transfer learning and multilingual pre-training can deliver competitive performance even for code-mixed text; this is an approach directly applicable to languages with high social media usage and code-switching patterns. \\

\noindent
Moreover, from a methodological perspective, techniques such as transfer learning from high-resource to low-resource contexts \cite{dione2023_masakha_pos, ogundepo-etal-2023-cross, beukman2023_ner, oladipo2023_better, onifade2018_embedded_fuzzy_dictionary,olalekan2022_machine_learning_for_naija, adebara2023_serengeti, gehrmann2023_tata, ruder2023_xtreme-up}, multilingual pre-training with mBERT, XLM-R, among others  \cite{adelani2021_masakhaner_1, dione2023_masakha_pos, beukman2023_ner, oladipo2023_better}, and cross-lingual embeddings as in MasakhaNER \cite{adelani2021_masakhaner_1} represent broadly applicable strategies. These methods allow resource-constrained languages to leverage shared morphosyntactic or semantic structures in related languages. Furthermore, community-driven speech corpus development initiatives such as \`{I}r\`{o}y\`{i}nSpeech \cite{ogunremi2024_iroyinspeech} demonstrate how volunteer-based, open-licence data collection can bootstrap Automatic Speech Recognition (ASR) and Text-to-Speech (TTS) systems for other low-resource languages with oral traditions.\\

\noindent
Furthermore, socio-cultural factors identified in Yor\`{u}b\'{a} NLP, such as code-switching \cite{muhammad2022_naijasenti}, \cite{adewole2020_vowel_elision}, domain-specific vocabulary gaps \cite{liu2018_context_OOV}, \cite{akinade2014_number-to-text}, and reduced digital use \cite{developing_yoruba_corpus}, \cite{Asubiaro2021ASR}, are also common in many other low-resource languages. Addressing these requires culturally informed NLP design, specifically incorporating language-specific orthography. In addition, as shown in Yor\`{u}b\'{a} NLP, integrating such considerations into dataset curation and model fine-tuning not only improves performance but also enhances user adoption and long-term language dynamism. Although cultural and societal barriers, such as the preference for English in formal language usage, might superficially be interpreted as a decline in relevance, it is logically seen instead as evidence of a critical gap in technological inclusion, based on available research \cite{abdulkareem2016_yorcall}.  Yor\`{u}b\'{a} language remains a vibrant, widely spoken language, and its strong cultural presence highlights that the challenge is not diminishing demand, but rather insufficient technological support. Thus, the limitation of NLP tools for Yor\`{u}b\'{a} should be understood as a call for urgent resource development, ensuring the language remains visible and usable in digital technology, rather than as an indicator of declining societal importance.\\

\noindent
In summary, the Yor\`{u}b\'{a} NLP landscape offers a transferable blueprint for advancing NLP in other low-resource languages through the evident foundational resources, application of hybrid methodologies, adopting multi-lingual pre-training and cross-lingual transfer, thereby accelerating progress, and the incorporation of socio-cultural knowledge into the technology lifecycle. These principles, grounded in the evidence presented in Tables \ref{tab5} - \ref{tab7}, provide a replicable framework for fostering inclusive and sustainable NLP development across diverse low-resource language contexts. By considering the case of the Yor\`{u}b\'{a} language within a broader low-resource languages landscape, this discussion emphasizes the potential for synergy of methods, tools, and collaborative frameworks, thereby helping to ensure that research findings from one language can be effectively leveraged to support many others.

\subsection{Current Yor\`{u}b\'{a} NLP in Relation to State-of-the-Art Practices}

Existing studies in Yor\`{u}b\'{a} NLP reveal both progress and significant gaps when compared to mainstream NLP research. In terms of task coverage, work has primarily focused on foundational areas such as part-of-speech tagging, machine translation, sentiment analysis, and limited speech recognition. However, several high-impact tasks, including question answering, dialogue systems, summarization, sarcasm detection, identification of harmful language usage, including abusive, offensive, hate speech, or cyberbullying, and also multimodal applications, remain largely unexplored. \\

\noindent
Methodologically, Yorùbá NLP has mostly relied on rule-based systems and statistical and traditional machine learning approaches \cite{yusuf_hierarchical, olalekan2022_machine_learning_for_naija, Abiola_hidden_markov}, with limited research tasks leveraging transformer-based architectures and PLMs. In contrast, state-of-the-art NLP is dominated by large-scale pretrained models like Grok, GPT, LLaMA, Qwen models, among others, and task-specific fine-tuning, which enable broad generalization and high accuracy. The relative scarcity of Yor\`{u}b\'{a}-specific training data in these domains has limited the application of these methods.\\

\noindent
Furthermore, while some corpora, lexicons, and parallel datasets exist as presented in Table \ref{tab7}, they are often fragmented or domain-limited. Unlike high-resource languages, Yor\`{u}b\'{a} NLP lacks many benchmark datasets and evaluation suites that facilitate reproducibility and comparison across approaches. \\

\noindent
Altogether, these gaps highlight clear priorities for future research, by looking into the current reality of NLP involving Yor\`{u}b\'{a}, and the general state-of-the-art. Efforts should focus on developing benchmark datasets for core tasks, adopting more advanced transfer learning and multilingual fine-tuning strategies, and expanding into overlooked task areas such as dialogue systems and multimodal NLP.

\subsection{Limitations of Study}
\noindent
While this systematic literature review investigates the progress and status of NLP involving Yor\`{u}b\'{a}, it is noteworthy to mention that ``Yor\`{u}b\'{a}'' in this case is not specific to a certain dialect of the language, such as Yor\`{u}b\'{a} of  If\d{\`{e}}, \`{I}j\d{\`{e}}b\'{u}, or \`{I}l\`{a}j\d{e}. The study recognised Yor\`{u}b\'{a} language as one encompassing several dialects across different countries. Even though certain sections of it specifically highlighted NLP research and resources involving Yor\`{u}b\'{a} dialects \cite{ahia2024voices_unheard}, this study might not be sufficient when dialects of the language are the sole focus in the NLP research. \\

\noindent
Furthermore, the last date for retrieving information for the study was October $2024$. Hence, publications emerging afterwards would not have been included in the synthesis. Similarly, only peer-reviewed studies were included to ensure high quality and reliability in findings. This might limit a recently published relevant study undergoing a peer-review process during the period the databases were searched. Nevertheless, the systematic review ensures a substantial representation of information from various primary studies by considering a decade of publication years.

\section{Conclusion and Future Directions}
This section summarises the study and process involved and aims to inform readers of possible research areas towards improving Yor\`{u}b\'{a} language involvement in NLP research.
\subsection{Conclusion}
This SLR explores NLP progress involving Yor\`{u}b\'{a} language. It involves surveying studies between $2014$ and $2024$, which have used Yor\`{u}b\'{a} language in their NLP research, and those with great emphasis on the language, in case of a multilingual setting. Data were synthesized from these primary studies to deduce findings, thereby providing answers to the established four research questions, forming the research's core objectives. \\

\noindent
The research questions explore the tasks, techniques, language resources, and challenges involved in Yor\`{u}b\'{a} NLP over a decade. Established protocols and guidelines were followed systematically to ensure maximum formal inundation, eliminating possible bias. Moreover, the information synthesised from the $105$ primary studies has been carefully reported, encapsulating the relevant findings from the systematic review. \\

\noindent
In conclusion, with this study, language researchers will be abreast of the current progress in Yor\`{u}b\'{a} NLP, thereby equipping them with the necessary ideas to preserve the language through NLP tool representation---this is crucial as it is a widely spoken language with abundant cultural richness and linguistic features.  Similarly, it is required to guide researchers taking this route in the future to eliminate or limit possible odysseys in their research journey.

\subsection{Future Directions}
Even though the findings show promising results for NLP research involving  Yor\`{u}b\'{a} language, it is pertinent to outline the current absence of significant efforts in some domains, which are equally important towards improving  NLP in the language.  For instance, research efforts involving the identification of abusive, offensive, hate speech, or cyberbullying, which are all regarded as harmful language usage, have limited research as of the time of this publication. This could be due to a lack of or limited annotated corpus in this domain.  Such research tasks are essential for ameliorating the usage rate of sensitive words, phrases, and sentences on social media, as they could potentially endanger other users. Consequently, future research efforts can be directed toward building relevant corpora to address these NLP challenges and creating relevant benchmarks to facilitate empirical analysis and continuous research. \\

\noindent
Sarcasm detection is another that remains untapped yet for  Yor\`{u}b\'{a} and most under-resourced languages, as noted in the critical analysis compared with the state-of-the-art. Sarcasm is used to mask the true emotion in a state, mostly by exuding positivity or a seeming positive demeanour. This phenomenon of saying what is not meant or meaning what is not said by natural language users poses challenges for NLP. However, it is essential to accurately detect users' real emotions for different purposes, including product reviews, feedback analysis, politics \& governance, among others. Consequently, developing relevant sarcasm corpora and benchmarks that will potentially birth advanced sarcasm detection models is crucial. Future research can be focused on this domain to promote certainty in natural language usage.\\

\noindent
Moreover, general resource development tasks are essential to limit the paucity of Yor\`{u}b\'{a} language resources. This could be achieved through continuous collaboration, open-source, and community initiatives, such as the one carried out through Maskhane\footnote{\url{https://www.masakhane.io/}}. Moreover, more priority can be given to developing Yor\`{u}b\'{a}-specific pre-trained models and fine-tuning existing multilingual models for better performance in under-resourced settings. \\

\noindent
Furthermore, more research should be directed towards solving linguistic challenges such as tonal variations, morphology complexities, and diacritic restoration, as these are essential for decoding the nuances in the language.  Developing pre-processing tools and models that cater to and integrate linguistic knowledge or incorporate phonological features in the language could significantly improve performance. \\

\noindent
Finally, addressing the drawbacks associated with cultural and social challenges requires developing context-aware models that can adapt to real-world changes. Continuous language usage would also benefit from exploring new cases, such as LLM conversational agents, and developing healthcare and educational tools incorporating NLP. These innovations can aid in bridging the digital gaps among the Yor\`{u}b\'{a}-speaking communities. \\

\noindent
\textbf{Declaration of Conflict of Interest}\\
The authors attest that no conflict of interest is involved in the publication. \\

\newpage
\noindent
\textbf{Acknowledgements}\\
This publication emanated from research conducted with the financial support of Science Foundation Ireland under Grant number 18/CRT/6049. For Open Access, the author has applied a CC BY public copyright licence to any Author Accepted Manuscript version arising from this submission. The authors also show gratitude to the reviewers for their painstaking efforts. \\

\noindent
\textbf{Data Availability}\\
No data was used for this research task. However, datasheets that record the different stages in the SLR can be found in the GitHub repository\footnote{\url{https://github.com/toheebadura/SLR}} for reference. Also, the published paper can be found online on ScienceDirect\footnote{\url{https://doi.org/10.1016/j.nlp.2025.100194}}.

\renewcommand{\bibname}{the_references}
\bibliographystyle{unsrt}
\bibliography{the_references}
\addcontentsline{toc}{chapter}{References}
\end{document}